%% file: eccv.tex
\documentclass[runningheads]{llncs}

\usepackage{eccv}

\usepackage{eccvabbrv}

\usepackage{graphicx}
\usepackage{booktabs}
\usepackage{multirow}
\usepackage{amsmath}    % For align environment and equations
\usepackage{amssymb}    % For \mathbb (e.g., \mathbb{R})
\usepackage{bm}         % Optional: for bold math symbols (if needed)
\usepackage{multicol}
\usepackage{subcaption}
\usepackage{arydshln}
\usepackage{placeins}
\usepackage{multirow}
\usepackage{nicefrac}
\usepackage{xcolor}
\usepackage{xspace}
\newcommand{\method}{CLIMP\xspace}

\usepackage[accsupp]{axessibility}  % Improves PDF readability for those with disabilities.

\usepackage{hyperref}

\usepackage{orcidlink}

\begin{document}
% ---------------------------------------------------------------
\title{\method: \\ Contrastive Language-Image Mamba Pretraining}

\titlerunning{CLIMP: Contrastive Language-Image Mamba Pretraining}

\author{Nimrod Shabtay\inst{1,2} \and Itamar Zimerman\inst{1} \and Eli Schwartz\inst{2} \and Raja Giryes\inst{1}}

\authorrunning{N.~Shabtay et al.}

\institute{Tel-Aviv University \and IBM Research}

\maketitle

\input{figures/block_diagram}

\input{abstract}
\input{introduction}

\input{rw}
\input{method}
\input{results}

\input{conclusions_limitations}
\clearpage
\bibliographystyle{splncs04}
\bibliography{custom}

%\clearpage
% \onecolumn
%\appendix
%\input{appendix}

\end{document}

% ---------------------------------------------------------------
\title{\method: \\ Contrastive Language-Image Mamba Pretraining}

\titlerunning{\method: Contrastive Language-Image Mamba Pretraining}

\author{Nimrod Shabtay\inst{1,2} \and Itamar Zimerman\inst{1} \and Eli Schwartz\inst{2} \and Raja Giryes\inst{1}}

\authorrunning{N.~Shabtay et al.}

\institute{Tel-Aviv University \and IBM Research}

\maketitle
\vspace{-15pt}
\section{CLIP Benchmarks - Full results}
\vspace{-10pt}
We report detailed per-dataset breakdowns of the zero-shot classification results from Table 1 (main paper) in Tables~\ref{tab:retrieval} and \ref{tab:zeroshot_classification_full}.
\begin{table}[hbtp!]
\vspace{-10pt}
\centering
\resizebox{\textwidth}{!}{%
\begin{tabular}{l|cc|cc|cc|cc|cc|cc}
\toprule
& \multicolumn{2}{c|}{ViT-B-P-16+LLaMA} & \multicolumn{2}{c|}{VMamba-B+Mamba2} & \multicolumn{2}{c|}{VMamba-B+Mamba1} & \multicolumn{2}{c|}{RoPE-ViT-B+LLaMA} & \multicolumn{2}{c|}{Flex-ViT-B+LLaMA} & \multicolumn{2}{c}{NaFlex-ViT-B+LLaMA} \\
Dataset & T-R@5 & I-R@5 & T-R@5 & I-R@5 & T-R@5 & I-R@5 & T-R@5 & I-R@5 & T-R@5 & I-R@5 & T-R@5 & I-R@5 \\
\midrule
flickr30k & 82.7 & 72.9 & 85.0 & 76.9 & 87.5 & 76.5 & 82.9 & 73.9 & 81.9 & 72.8 & 81.1 & 72.1 \\
flickr8k & 78.4 & 68.1 & 79.3 & 69.1 & 81.7 & 70.5 & 77.6 & 68.1 & 77.0 & 66.6 & 81.5 & 66.0 \\
mscoco\_captions & 57.5 & 47.3 & 61.4 & 49.5 & 61.9 & 49.6 & 58.3 & 48.1 & 58.1 & 47.7 & 57.1 & 46.3 \\
\bottomrule
\end{tabular}%
}
\caption{Zero-shot Retrieval Results. We report Text Retrieval Recall@5 (T-R@5) and Image Retrieval Recall@5 (I-R@5) (\%) on standard retrieval benchmarks.}
\vspace{-35pt}
\label{tab:retrieval}
\end{table}
 \vspace{-10pt}
 \section{Image-Text Similarity additional Visualizations}
\begin{figure}[h!]
\vspace{-10pt}
    \centering
    \setlength{\tabcolsep}{1pt}
    \newcommand{\imgbox}[1]{\adjustbox{width=0.24\textwidth, height=0.18\textwidth, keepaspectratio, valign=c}{#1}}
    \begin{tabular}{cccc}             
        Model (similarity) & Flex-ViT (0.539) & RoPE-ViT (0.581) & \method (0.578) \\
        \imgbox{\includegraphics[width=\linewidth]{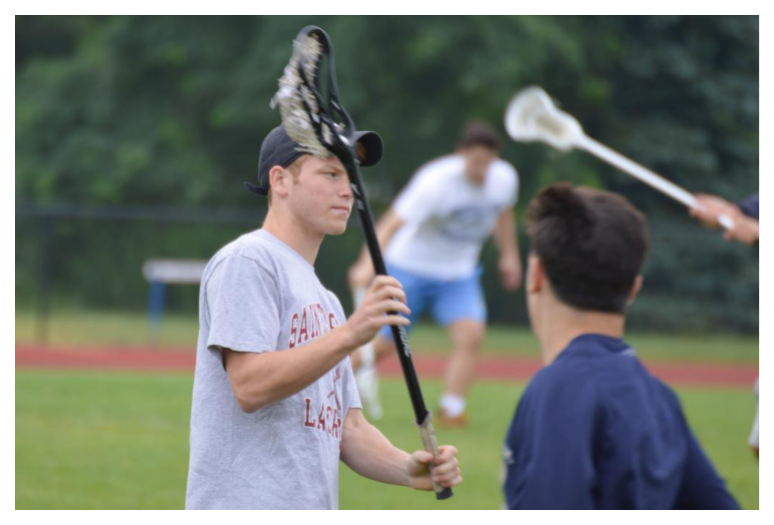}} &
        \imgbox{\includegraphics[width=\linewidth]{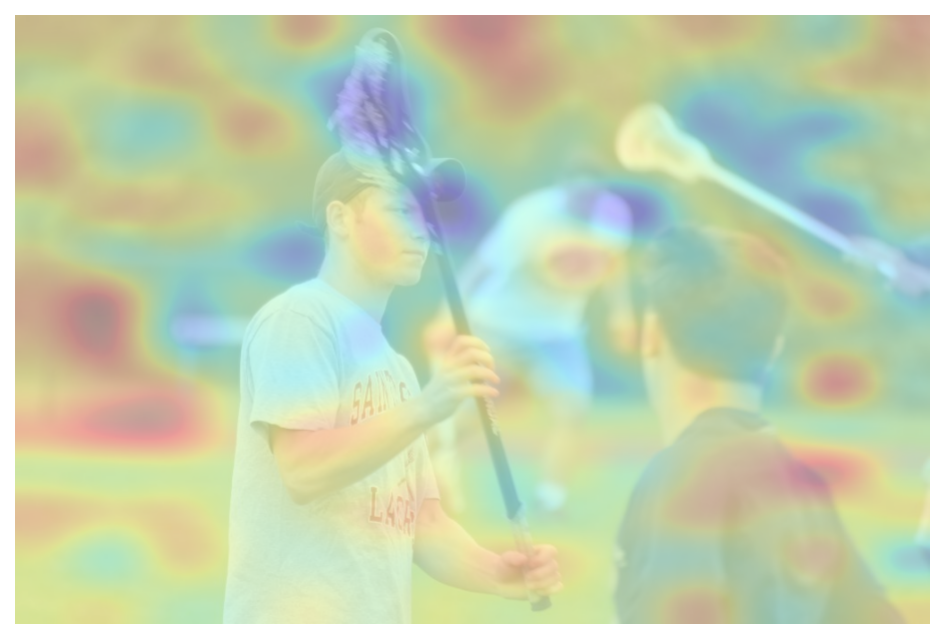}} &
        \imgbox{\includegraphics[width=\linewidth]{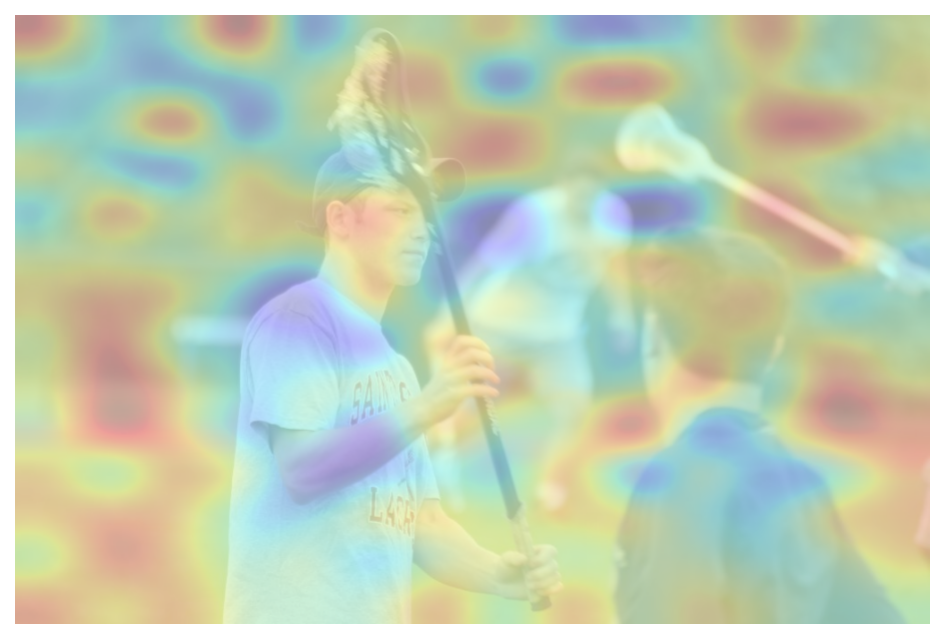}} &
        \imgbox{\includegraphics[width=\linewidth]{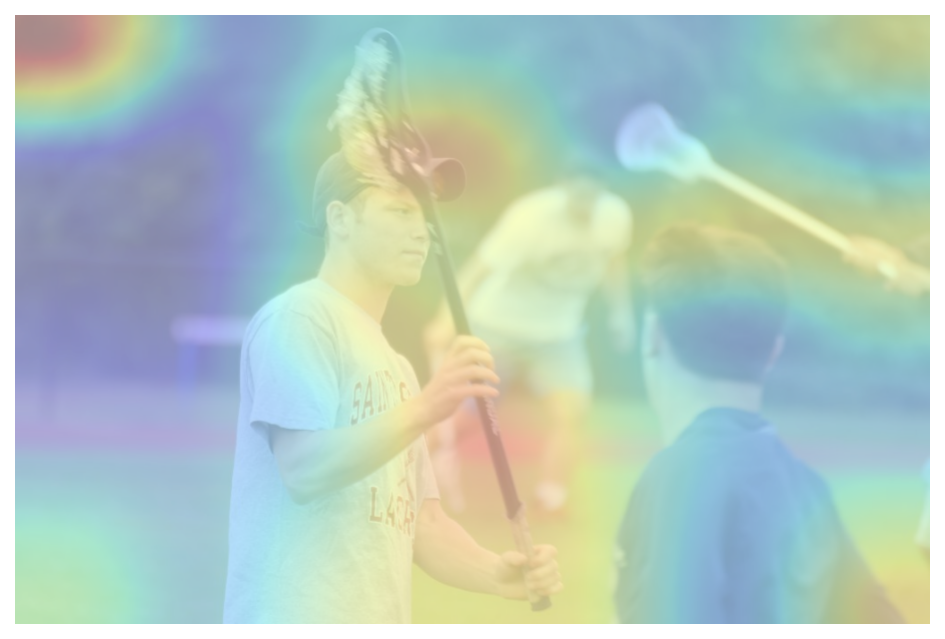}} \\[2pt]
        Model (similarity) & Flex-ViT (0.619) & RoPE-ViT (0.633) & \method (0.661) \\
        \imgbox{\includegraphics[width=\linewidth]{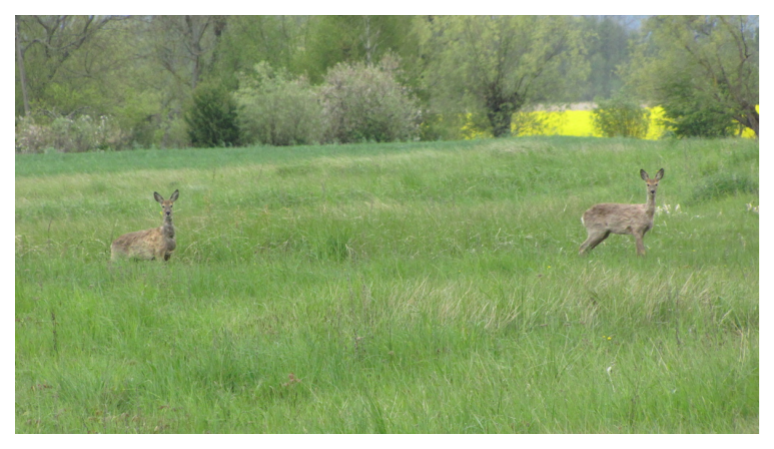}} &
        \imgbox{\includegraphics[width=\linewidth]{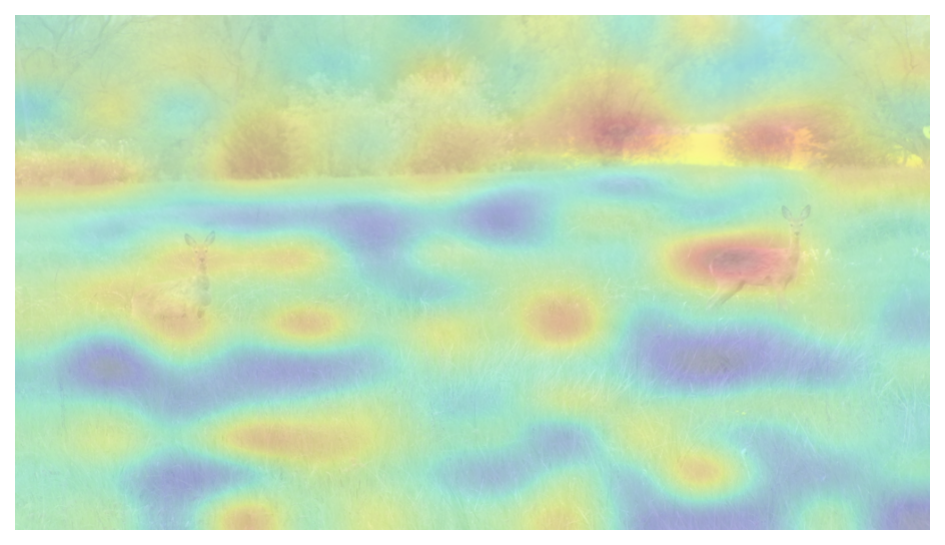}} &
        \imgbox{\includegraphics[width=\linewidth]{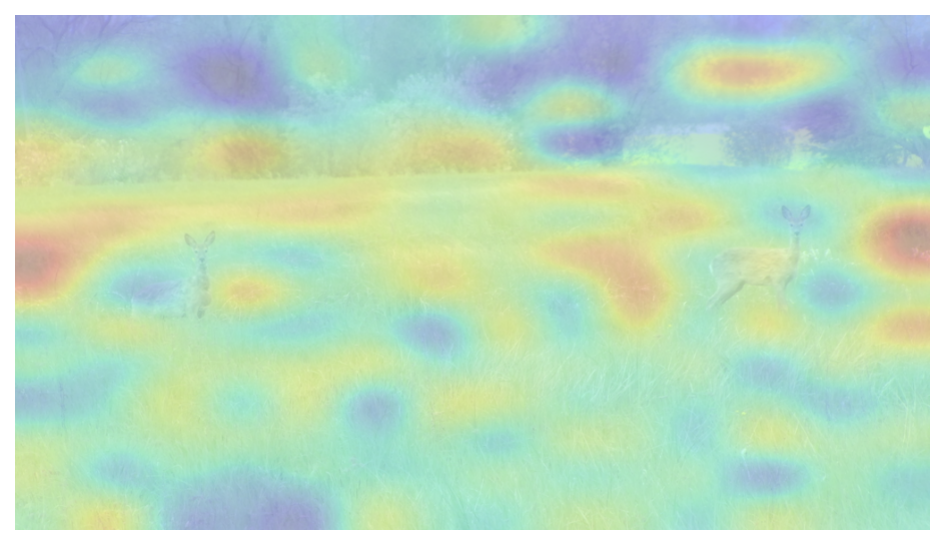}} &
        \imgbox{\includegraphics[width=\linewidth]{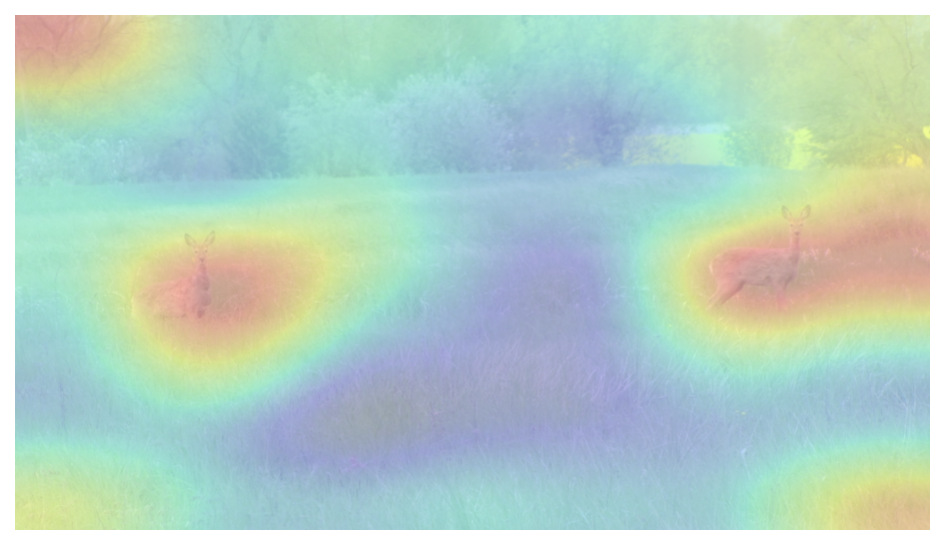}} \\[2pt]
        Model (similarity) & Flex-ViT (0.499) & RoPE-ViT (0.516) & \method (0.564) \\
        \imgbox{\includegraphics[width=\linewidth]{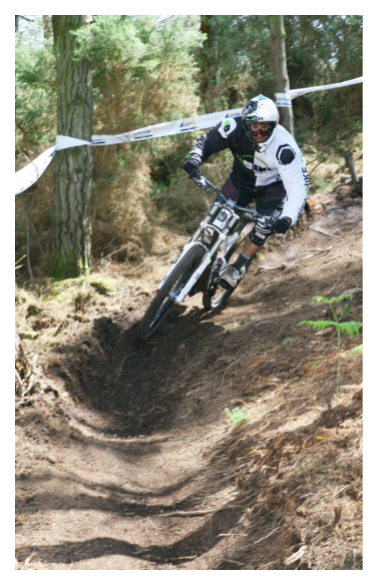}} &
        \imgbox{\includegraphics[width=\linewidth]{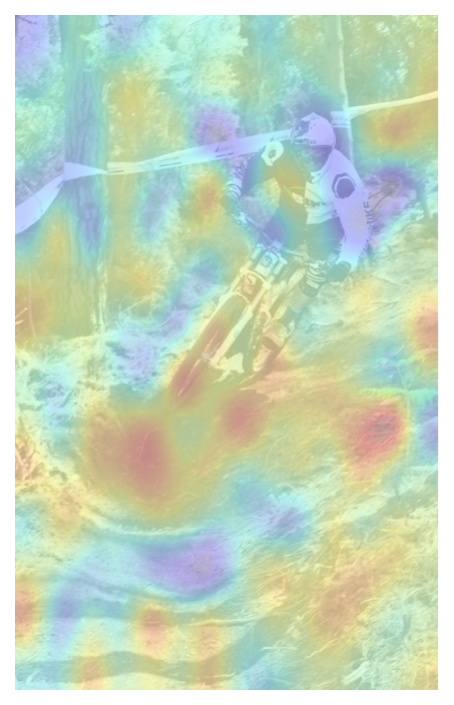}} &
        \imgbox{\includegraphics[width=\linewidth]{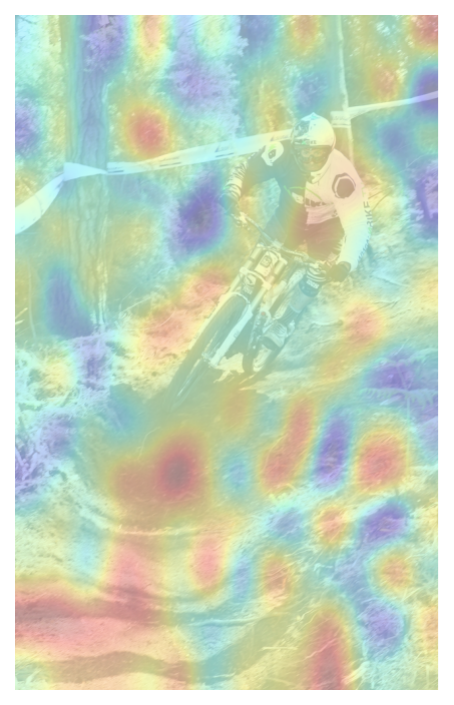}} &
        \imgbox{\includegraphics[width=\linewidth]{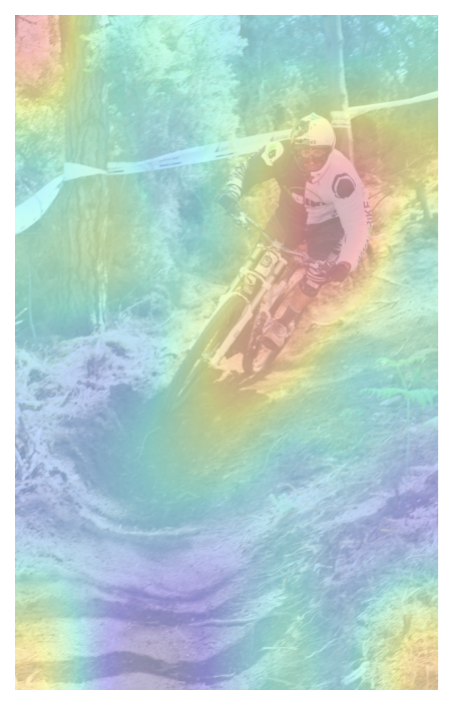}} \\
    \end{tabular}
    \caption{Visualization of Image-Text similarity for the captions: "A person is holding a lacrosse stick while a person watches." (top), ``Two deer are standing in a lush field of grass near a wooded area.'' (middle) and ``A man in a white and black uniform racing a bike in the dirt.'' (bottom)}
    \vspace{-20pt}
    \label{fig:nocaps_visuals_sup}    
\end{figure}

 \vspace{-10pt}
We provide additional image-text similarity heatmaps in Figure~\ref{fig:nocaps_visuals_sup}, comparing patterns across Flex-ViT, RoPE-ViT, and \method{} for diverse scene types. The Transformer-based variants (Flex-ViT and RoPE-ViT) exhibit diffuse attention patterns that rely on spurious correlations, whereas \method{} produces sharper, more interpretable patterns that better localize semantically relevant regions.

 \begin{table}[t!]                                                                    
  \centering
  \small                                                    
  \resizebox{\textwidth}{!}{%
  \begin{tabular}{llccccc}
  \toprule
  Dataset & Metric & FlexViT-B/16 & NaFlex-B/16 & RoPE-ViT-B/16 & \method (Mamba-2) & \method (Mamba-1) \\
  \midrule
  \multirow{2}{*}{ImageNet-1K} & Acc@1 & 38.3 & 36.6 & 40.7 & \underline{43.5} & \textbf{44.0} \\
   & Acc@5 & 69.0 & 67.0 & 71.5 & \underline{73.9} & \textbf{74.8} \\
  \multirow{2}{*}{ImageNet-V2} & Acc@1 & 32.8 & 31.4 & 34.4 & \underline{37.0} & \textbf{37.5} \\
   & Acc@5 & 62.5 & 60.5 & 65.4 & \underline{67.6} & \textbf{68.4} \\
  \multirow{2}{*}{ImageNet-A} & Acc@1 & 13.2 & 12.0 & \textbf{16.3} & 15.5 & \underline{15.6} \\
   & Acc@5 & 41.2 & 36.0 & \textbf{46.3} & \underline{45.5} & 45.3 \\
  \multirow{2}{*}{ImageNet-R} & Acc@1 & 45.4 & 44.8 & \textbf{47.8} & 46.2 & \underline{46.6} \\
   & Acc@5 & 74.3 & 72.5 & \textbf{76.0} & 74.4 & \underline{74.5} \\
  \multirow{2}{*}{ImageNet-O} & Acc@1 & 38.0 & 38.5 & 40.1 & \underline{48.1} & \textbf{49.8} \\
   & Acc@5 & 68.2 & 67.6 & 69.8 & \underline{77.0} & \textbf{78.8} \\
  \multirow{2}{*}{ImageNet-Sketch} & Acc@1 & 24.6 & 23.5 & \underline{27.4} & 27.0 & \textbf{27.5} \\
   & Acc@5 & 50.0 & 49.1 & 53.1 & \underline{54.2} & \textbf{54.4} \\
  \multirow{2}{*}{ObjectNet} & Acc@1 & 28.1 & 26.8 & \textbf{33.1} & \underline{29.2} & 28.1 \\
   & Acc@5 & 54.3 & 53.4 & \textbf{60.9} & \underline{55.9} & 54.3 \\
  \multirow{2}{*}{Cars} & Acc@1 & 8.3 & 8.5 & 8.4 & \textbf{9.9} & \underline{9.1} \\
   & Acc@5 & 31.0 & 30.4 & 31.1 & \textbf{34.1} & \underline{32.4} \\
  \multirow{2}{*}{FGVC Aircraft} & Acc@1 & \textbf{2.6} & 1.9 & 1.7 & 2.4 & \underline{2.5} \\
   & Acc@5 & 9.4 & 8.3 & 8.6 & \textbf{9.9} & \underline{9.7} \\
  \multirow{2}{*}{Country211} & Acc@1 & 4.1 & 4.2 & 3.9 & \textbf{4.7} & \underline{4.3} \\
   & Acc@5 & 14.9 & 14.7 & 14.5 & \textbf{15.8} & \underline{15.4} \\
  \multirow{2}{*}{SUN397} & Acc@1 & \underline{49.1} & 48.8 & 47.5 & \textbf{49.5} & 48.2 \\
   & Acc@5 & 83.7 & 83.6 & 83.2 & \textbf{84.6} & \underline{84.2} \\
  \multirow{2}{*}{FER2013} & Acc@1 & 15.1 & 15.2 & \textbf{29.1} & \underline{22.2} & 18.8 \\
   & Acc@5 & 77.2 & 76.6 & \textbf{86.9} & \underline{83.1} & 81.7 \\
  \multirow{2}{*}{VOC2007} & Acc@1 & 65.4 & 68.3 & \underline{71.2} & 69.4 & \textbf{75.8} \\
   & Acc@5 & 88.9 & 91.5 & 92.2 & \underline{93.9} & \textbf{96.2} \\
  VOC2007 Multilabel & mAP & 81.1 & 80.1 & \underline{82.0} & 81.0 & \textbf{82.2} \\
  RenderedSST2 & Acc@1 & \textbf{50.4} & \underline{50.1} & 50.1 & 49.9 & 50.1 \\
  \multirow{2}{*}{CIFAR-10} & Acc@1 & 83.1 & 83.1 & \textbf{88.4} & \underline{84.8} & 83.0 \\
   & Acc@5 & 98.4 & 98.9 & 99.4 & \textbf{99.5} & \underline{99.4} \\
  \multirow{2}{*}{CIFAR-100} & Acc@1 & \underline{54.3} & 51.4 & \textbf{54.5} & 50.7 & 48.3 \\
   & Acc@5 & \underline{83.2} & 82.1 & \textbf{84.4} & 82.8 & 81.1 \\
  \multirow{2}{*}{SVHN} & Acc@1 & \textbf{20.8} & 9.4 & 7.0 & 9.8 & \underline{9.9} \\
   & Acc@5 & \textbf{62.9} & 55.4 & \underline{57.7} & 48.1 & 52.3 \\
  \multirow{2}{*}{CLEVR Count} & Acc@1 & 13.8 & 14.7 & \textbf{17.1} & 13.1 & \underline{16.2} \\
   & Acc@5 & 63.4 & 64.7 & \underline{70.4} & \textbf{71.0} & 63.8 \\
  \multirow{2}{*}{CLEVR Distance} & Acc@1 & 15.9 & 16.1 & 16.4 & \underline{16.8} & \textbf{18.3} \\
   & Acc@5 & 91.5 & \underline{91.9} & \textbf{93.9} & 91.9 & 90.9 \\
  \multirow{2}{*}{Diabetic Ret.} & Acc@1 & 12.7 & \underline{30.7} & 3.5 & 10.4 & \textbf{71.3} \\
   & Acc@5 & \textbf{100.0} & \underline{100.0} & 100.0 & 100.0 & 100.0 \\
  \multirow{2}{*}{DMLab} & Acc@1 & \textbf{21.2} & 11.9 & \underline{20.4} & 15.2 & 17.8 \\
   & Acc@5 & \underline{84.8} & 81.4 & 82.1 & \textbf{85.1} & 81.2 \\
  PatchCamelyon & Acc@1 & 49.4 & 50.6 & \textbf{52.1} & 46.6 & \underline{51.5} \\
  \multirow{2}{*}{dSprites Orient.} & Acc@1 & 2.7 & 2.1 & 2.6 & \textbf{2.8} & \underline{2.8} \\
   & Acc@5 & 13.0 & 11.8 & \underline{13.3} & \textbf{13.7} & 11.5 \\
  \multirow{2}{*}{dSprites X-Pos} & Acc@1 & 3.1 & 3.1 & \underline{3.1} & 2.9 & \textbf{3.2} \\
   & Acc@5 & 15.3 & \underline{15.5} & 15.4 & 15.2 & \textbf{15.6} \\
  \multirow{2}{*}{dSprites Y-Pos} & Acc@1 & 2.9 & \underline{3.2} & 3.1 & 3.1 & \textbf{3.4} \\
   & Acc@5 & 14.3 & 15.6 & 15.0 & \underline{15.8} & \textbf{16.2} \\
  \multirow{2}{*}{sNORB Azimuth} & Acc@1 & 5.1 & 5.1 & \underline{5.4} & \textbf{5.7} & 5.1 \\
   & Acc@5 & 27.6 & \textbf{28.6} & \underline{28.0} & 27.4 & 27.0 \\
  \multirow{2}{*}{sNORB Elevation} & Acc@1 & 9.9 & \textbf{12.0} & \underline{10.9} & 10.8 & 10.3 \\
   & Acc@5 & 54.7 & \textbf{58.5} & 54.7 & \underline{55.6} & 54.2 \\
  \midrule
  \textbf{Average} & Acc@1 & 26.3 & 26.1 & \underline{27.3} & 26.9 & \textbf{29.6} \\
  \textbf{Average} & Acc@5 & 57.3 & 56.6 & \underline{59.0} & \textbf{59.0} & 58.5 \\
  \bottomrule
  \end{tabular}%
  }
  \caption{\textbf{Zero-shot Classification Results.} Per-dataset Acc@1 and Acc@5 on 28 classification benchmarks (mAP for VOC2007 Multilabel). Best in
  \textbf{bold}, second best \underline{underlined}.}
  \label{tab:zeroshot_classification_full}
  \end{table}

% \begin{table}[h]
% \centering
% \setlength{\tabcolsep}{5pt}
% \small
% \resizebox{\linewidth}{!}{
% \begin{tabular}{lcccc}
% \toprule
%  & Best of ResNet & Best of ViT-B & \multicolumn{2}{c}{CLIMP} \\
% \cmidrule(lr){2-2} \cmidrule(lr){3-3} \cmidrule(lr){4-5}
% Task & RN50q-cc & ViT-cpC & Mamba-2 & Mamba-1 \\
% \midrule
% Classification & 31.93 & 17.77 & \underline{34.48} & \textbf{36.06} \\
% Retrieval      & 55.31 & 18.57 & \underline{58.81} & \textbf{59.67} \\
% \bottomrule
% \end{tabular}}
% \caption{OpenCLIP Results. We report the best of each model family in OpenCLIP. RN50q-cc refers to ResNet50-quickgelu trained on CC12M and ViT-cpC refers to ViT-B trained on CommonPool-S CLIP (13M).}
% \label{tab:open_clip}
% \end{table}

\begin{table}[h]
\centering
\setlength{\tabcolsep}{5pt}
\small
\resizebox{\linewidth}{!}{
\begin{tabular}{llccc}
\toprule
Family & Model & Training Size & Classification & Retrieval \\
\midrule
\multirow{4}{*}{ResNet}
 & RN50 & 12M & 31.80 & 54.75 \\
 & RN50-quickgelu & 12M & 31.93 & 55.31 \\
 & RN50 & 15M & 27.84 & 47.09 \\
 & RN50-quickgelu & 15M & 27.47 & 47.16 \\
\midrule
\multirow{7}{*}{ViT-B-32}
 & DataComp-S & 13M & 14.92 & 15.36 \\
 & CommonPool-S Basic & 13M & 14.45 & 18.74 \\
 & CommonPool-S CLIP & 13M & 17.78 & 18.57 \\
 & CommonPool-S Image & 13M & 14.92 & 15.36 \\
 & CommonPool-S LAION & 13M & 13.67 & 15.27 \\
 & CommonPool-S & 13M & 14.41 & 17.89 \\
 & CommonPool-S Text & 13M & 16.01 & 19.22 \\
\midrule                                                                                     \multirow{2}{*}{CLIMP}
 & Mamba-2 & 12M & \underline{34.48} & \underline{58.81} \\                                  & Mamba-1 & 12M & \textbf{36.06} & \textbf{59.67} \\
\bottomrule                                                                                  \end{tabular}}
\caption{OpenCLIP benchmark results for all model variants. Classification is average acc@1 over 35 datasets. Retrieval is the average of 19 metrics (R@1/R@5/R@10 on Flickr30k and MSCOCO, plus WinoGAViL jaccard scores)}
\label{tab:open_clip_full}
\end{table}     

\begin{table}[h]
\centering
\begin{tabular}{ccc}
\includegraphics[width=0.3\linewidth,height=0.3\linewidth]{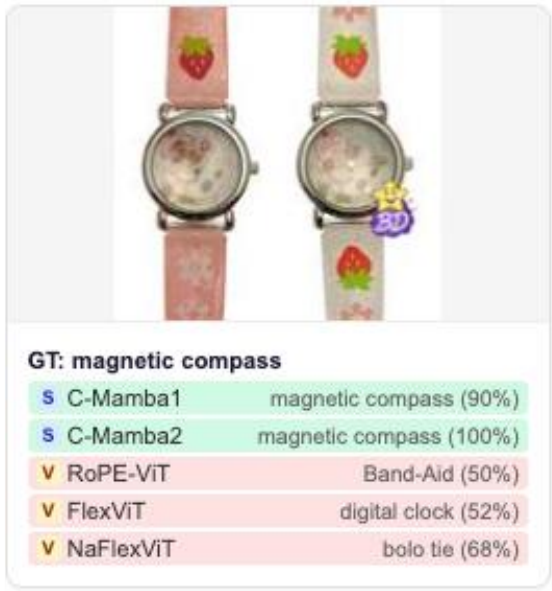} &
\includegraphics[width=0.3\linewidth,height=0.3\linewidth]{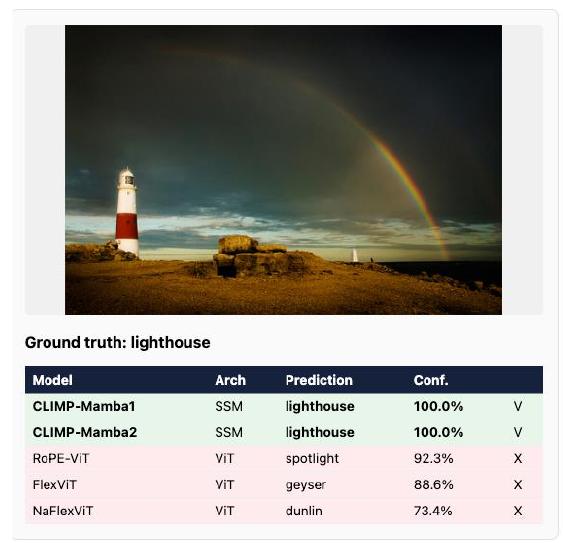} &
\includegraphics[width=0.3\linewidth,height=0.3\linewidth]{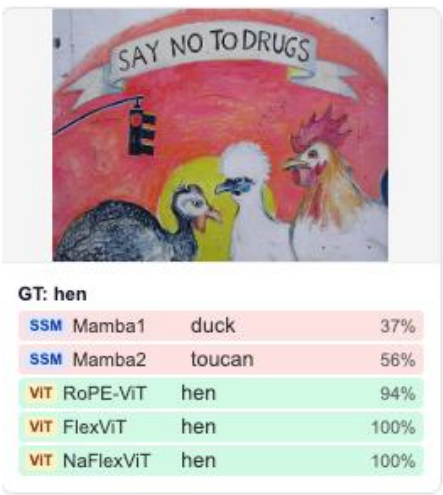} \\
\end{tabular}
\caption{We visualize favorable properties for when \method{} is favorable and when Transformer based CLIP. \method{} is better on natural images and Out-Of-Distribution as can seen on the left and center examples while Transformers are favorable on sketch / drawing visuals (right image).}
\label{tab:visual_favorable}
\end{table}

%% file: figures/block_diagram.tex
\begin{figure*}[t]
\centering
%\resizebox{0.65\columnwidth}{!}{%
\vspace{-10pt}
\includegraphics[width=0.74\linewidth]{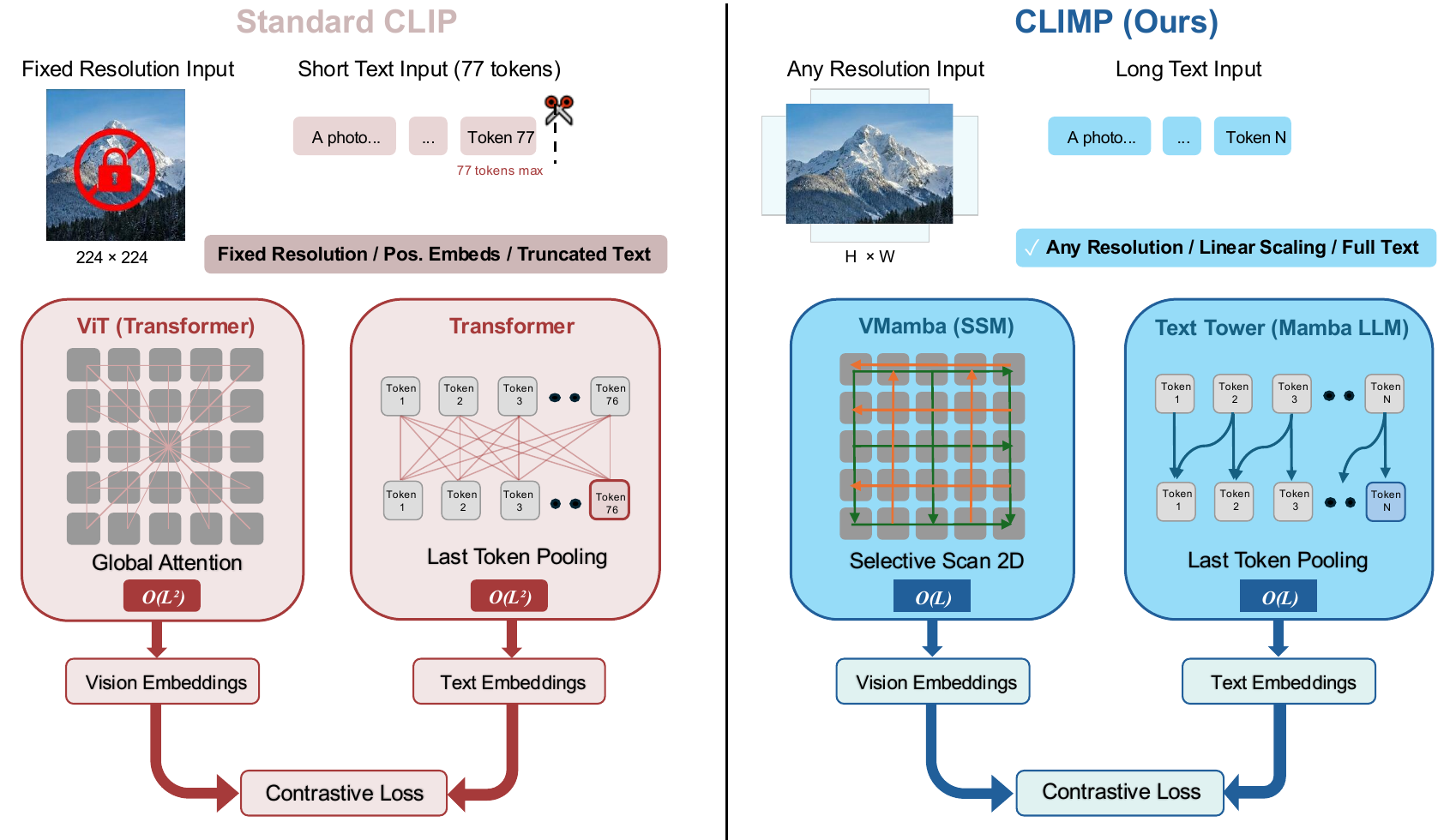}
% \vspace{-29pt}
\caption{\textbf{CLIP vs. CLIMP.} By replacing Transformer encoders with Mamba-based models, CLIMP achieves sub-quadratic $O(L)$ complexity instead of quadratic $O(L^2)$. It also removes the fixed resolution and 77-token text limitations of standard CLIP.}
\label{fig:block_diagram}
\vspace{-12pt}
\end{figure*}

%% file: abstract.tex
\begin{abstract}
Contrastive Language-Image Pre-training (CLIP) relies on Vision Transformers whose attention mechanism is susceptible to spurious correlations and scales quadratically with resolution.
To address these limitations, we present \method, the first fully Mamba-based contrastive vision-language model that replaces both the vision and text encoders with state-space architectures.
VMamba's cross-scan mechanism captures spatial inductive biases that reduce reliance on spurious correlations, producing an embedding space with tighter cross-modal alignment and lower hubness - geometric properties that translate to superior retrieval and out-of-distribution robustness, surpassing even CLIP-ViT-B trained on a dataset 167$\times$ larger on ImageNet-O.
\method naturally supports variable input resolutions without positional encoding interpolation or specialized training, achieving up to 6.6\% higher retrieval accuracy at 16$\times$ training resolution while using 5$\times$ less memory and 1.8$\times$ fewer FLOPs.
Mamba's autoregressive nature further enables processing of arbitrarily long text, overcoming CLIP's fixed 77-token context limitation for dense captioning retrieval.
Our scaling experiments across model sizes and dataset sizes show consistent, unsaturated improvements - indicating that \method's architectural advantages are not limited by training scale.
These results demonstrate that Mamba is a compelling alternative to Transformers for vision-language pre-training.
The code and models are publicly available at \href{https://github.com/NimrodShabtay/CLIMP}{https://github.com/NimrodShabtay/CLIMP}
\end{abstract}

%% file: introduction.tex
\section{Introduction}
\label{sec:introduction}

Contrastive Language-Image Pre-training (CLIP) \cite{radford2021learning} is a fundamental approach for learning transferable visual representations through natural language supervision. By aligning image and text embeddings in a shared latent space, CLIP enables zero-shot transfer across a range of downstream tasks. However, the Vision Transformer (ViT)~\cite{dosovitskiy2021image} backbone commonly employed in CLIP exhibits quadratic computational complexity with respect to sequence length, posing challenges when processing high-resolution images. Moreover, the pairwise token interactions in self-attention can be susceptible to spurious correlations~\cite{zhou2025fighting, tamayo2025spuriousvit}, leading to fragile representations under distribution shift.

Mamba~\cite{gu2024mamba}, a state-space model (SSM), has shown promising results as an alternative to Transformers in sequence modeling. It achieves sub-quadratic complexity in sequence length $L$ during training and constant-time complexity during inference, with these efficiency benefits becoming more pronounced in long-context scenarios. These properties make Mamba an attractive backbone for vision-language tasks requiring long-context processing such as dense captioning.

In the vision domain, adaptations such as Vision Mamba (Vim)~\cite{zhu2024vision}, Simba~\cite{patro2024simba} and VMamba~\cite{liu2024vmamba} have applied state space models to image understanding tasks with encouraging results. Beyond computational considerations, which allow high resolution processing, these models introduce spatial inductive biases that differ from the pairwise token interactions characteristic of self-attention. Prior work has shown that such spatial bias benefits vision tasks by enabling more robust~\cite{malik2025towards,du2024understanding} and sample-efficient learning~\cite{dascoli2021convit} - properties that ViTs lack.
However, the integration of Mamba-based vision encoders into contrastive vision-language frameworks remains relatively unexplored. Prior work~\cite{huang2024clipmamba} has investigated hybrid architectures combining Mamba and Transformer modules within individual towers, but does not explore a fully SSM-based dual-encoder architecture nor provide comprehensive evaluation across retrieval, robustness, resolution flexibility, and dense captioning. We empirically confirm that VMamba's spatial inductive bias transfers to the vision-language setting, enabling improved sample efficiency and robustness (Section~\ref{subsec::analysis}).

In this work, we present Contrastive Language-Image Mamba Pretraining (\method), a fully Mamba-based vision-language model that replaces both the vision and text encoders with state-space architectures within the CLIP framework. We pair VMamba~\cite{liu2024vmamba} as the vision encoder with Mamba-1~\cite{gu2024mamba} or Mamba-2~\cite{dao2024mamba2} language models as text encoders, creating the first end-to-end SSM-based contrastive vision-language model. Our controlled experiments on CC12M~\cite{cc12m} isolate architectural differences from data effects, and our scaling experiments across both model and dataset sizes demonstrate that \method's advantages are consistent and unsaturated, indicating they extend beyond our training regime.

\textbf{Our main contributions are:} %\noindent
\textbf{(1)} \method, the first fully Mamba-based CLIP model, integrating VMamba for vision and Mamba LLMs for text.
\textbf{(2)} Superior retrieval performance driven by tighter cross-modal alignment and reduced hubness in the embedding space.
\textbf{(3)} Strong out-of-distribution robustness, achieving the highest average accuracy across five ImageNet variants. On ImageNet-O, we surpass CLIP-ViT-B/16 trained on LAION-2B - a dataset 167$\times$ larger than ours - suggesting that architectural inductive biases can outweigh data scale for OOD robustness.
\textbf{(4)} Native variable-resolution support without complex positional encoding schemes or dedicated training, with significantly lower memory and compute overhead.
\textbf{(5)} Favorable scaling behavior across model sizes (22M--87M parameters) and dataset sizes (1M--12M pairs) with unsaturated performance curves, indicating that \method's advantages are architectural and should persist at larger scales.

%% file: rw.tex
% \vspace{-2pt}
\section{Related Work}
\vspace{-2pt}
\label{sec:related}

\noindent \textbf{Contrastive Vision-Language Learning.}
%Contrastive Language-Image Pre-training
CLIP~\cite{radford2021learning} marked a paradigm shift in vision-language learning by demonstrating that models trained on large-scale image-text pairs can achieve remarkable zero-shot transfer capabilities. %CLIP
It learns a joint embedding space where semantically similar images and texts are mapped close together, enabling open-vocabulary recognition without task-specific fine-tuning. Following CLIP's success, numerous works have sought to improve upon its framework. OpenCLIP~\cite{cherti2023reproducible} provides an open-source reproduction with reproducible scaling laws, training models on datasets such as LAION-400M~\cite{laion400m} and LAION-2B~\cite{schuhmann2022laion}. SigLIP~\cite{zhai2023sigmoid} replaces the softmax-based contrastive loss with a pairwise sigmoid loss, improving memory efficiency and enabling training with smaller batch sizes. EVA-CLIP~\cite{sun2023evaclip} enhances the training recipe with stronger vision and text encoders, achieving state-of-the-art zero-shot performance. ALIGN~\cite{ALIGN} scaled the training data with noisy image-text pairs. Recent work has also explored improving multimodal representations through caption adaptation and knowledge-enhanced supervision \cite{yanuka2025bridging}. More recent efforts include MetaCLIP~\cite{xu2024demystifying}, which introduces data curation algorithms for balanced training distributions, and SigLIP 2~\cite{tschannen2025siglip2}, which adds captioning-based pretraining and self-supervised losses. A recent work has also investigated improving the interpretability of vision-language models through structured semantic representations \cite{Yellinek20253VL}.

Despite these advances, all existing CLIP variants rely on Transformer-based vision encoders, predominantly the Vision Transformer (ViT)~\cite{dosovitskiy2021image}. While ViT has proven highly effective, its quadratic complexity with respect to sequence length poses challenges for high-resolution image processing. Scaling ViT to higher resolutions requires substantial computational resources and for a dynamic resolution processing ViT requires a specialized positional encoding schemes such as RoPE~\cite{su2024roformer} or dedicated training scheme~\cite{beyer2023flexvit}. Furthermore, studies have shown that CLIP's robustness to distribution shifts, while impressive on ImageNet variants~\cite{taori2020measuring,fang2022data}, may be overestimated when evaluated on datasets specifically designed to probe spurious correlations~\cite{wang2024sober}. These limitations motivate our exploration of alternative vision backbones that offer improved efficiency at high resolutions while enhancing robustness.

Recent work explored using LLMs as CLIP text encoders. LLM2CLIP~\cite{huang2024llm2clip} fine-tunes CLIP by replacing its text encoder with a Transformer-based LLM and training a lightweight adaptor, while UniViTAR~\cite{qiao2025univitar} pairs LLaMA with a ViT-based native-resolution framework. In contrast, \method uses Mamba LLMs as text encoders, enabling a fully SSM-based architecture with consistent sub-quadratic complexity across both modalities - rather than enhancing an existing Transformer-based CLIP or retaining a Transformer vision backbone.

\method is the first work to systematically investigate Mamba vision encoders within the CLIP framework. Replacing ViT with VMamba addresses the computational bottleneck of high-resolution processing and enables native variable-resolution inference that achieves improved zero-shot performance and robustness to distribution shifts.

\noindent \textbf{State Space Models (SSMs) for Vision.}
% SSMs have emerged as a promising alternative to Transformers for sequence modeling. Mamba \cite{gu2024mamba} introduced a selective mechanism that makes SSM parameters input-dependent, enabling content-aware reasoning with linear complexity. Mamba-2 \cite{dao2024mamba2} further refined this approach through the State Space Duality (SSD) framework, establishing theoretical connections between SSMs and attention while achieving 2-8$\times$ speedups.
SSMs have emerged as a promising alternative to Transformers for sequence modeling. Mamba~\cite{gu2024mamba} introduced a selective mechanism making SSM parameters input-dependent, enabling content-aware reasoning with linear complexity. Mamba-2~\cite{dao2024mamba2} refined this via the State Space Duality (SSD) framework, linking SSMs and attention while achieving 2-8$\times$ speedups.

Adapting Mamba for vision tasks presents unique challenges, as images are inherently 2D and lack the sequential structure of language. Vision Mamba (Vim)~\cite{zhu2024vision} addresses this by introducing bidirectional scanning with position embeddings, achieving competitive performance on ImageNet classification. 
SiMBA \cite{patro2024simba} addresses the stability issues of Mamba when scaling to large vision networks by introducing Einstein FFT for channel modeling, achieving state-of-the-art SSM performance on ImageNet and multiple time series benchmarks.
VMamba~\cite{liu2024vmamba} proposes the 2D Selective Scan (SS2D) module, which unfolds image patches along four traversal paths to capture global context while maintaining linear complexity. This cross-scan enables each patch to integrate information from all spatial directions, effectively establishing global receptive fields. Subsequent works have extended visual SSMs to various domains, including medical image segmentation~\cite{ruan2024vmunet}, video understanding~\cite{li2024videomamba}, and point cloud processing~\cite{liu2024point}.
Recent studies have also begun exploring the robustness properties of visual SSMs. %Du et al.~\cite{du2024understanding}
The works in \cite{du2024understanding,malik2025towards}
investigate the robustness of visual SSMs for image classification, finding that Mamba-based architectures exhibit different failure modes compared to ViTs under adversarial perturbations and distribution shifts. These findings suggest that the inductive biases of SSMs may offer complementary advantages to attention-based models in terms of robustness.

While prior work has established the effectiveness of visual SSMs for classification tasks, their potential for vision-language learning remains largely unexplored. \method bridges this gap by integrating VMamba into the CLIP framework, demonstrating that Mamba-based vision encoders can learn effective multimodal representations. Our experiments reveal that the architectural properties of SSMs - particularly their implicit handling of positional information through scanning patterns - enable native resolution flexibility and contribute to improved robustness on out-of-distribution benchmarks.

%% file: method.tex
\section{\method}
\label{sec:method}
\input{tables/main_results}

Mamba offers several advantages over attention models, including (1) improved long-context efficiency and scalability, (2) enhanced robustness, and (3) positional awareness. These long-context capabilities have recently been further theoretically \cite{bar2025revisiting} and empirically \cite{Ben-Kish2025DeciMamba,ben-kish2025overflow,lu2025mamba} analyzed proposing various techniques for improved length extrapolation and identifying mechanisms that stabilize recurrent state-space models during long-sequence processing. 

Given the above, replacing Transformers in CLIP entirely with Mamba can lead to sub-quadratic memory complexity in both modalities while maintaining improved representation quality.
To better understand the suitability of Mamba for CLIP, we discuss its \emph{inductive bias}, which is an important factor for sample-efficiency, out-of-distribution generalization, and representational capacity.

To empirically study this bias, we present positive results in Section~\ref{sec:inductive_bias} using synthetic tasks executed in controlled environments. Analytically, prior work has provided evidence that state-space layers exhibit inductive bias toward smoothness and locality compared to Transformers~\cite{zimerman2024viewing}. For Mamba-2, the inductive bias is easier to characterize: the recurrent update uses a per-step transition matrix $\bar{A}$ parameterized as $\bar{A} = \lambda I$, so under standard stability conditions, the influence of a state $k$ steps in the past decays as $\lambda^k$, encouraging locality and smoothness. For Mamba-1, the selective mechanism makes $\mathbf{B}$, $\mathbf{C}$, and $\Delta$ input-dependent, which precludes a simple closed-form characterization. However, our empirical results in Section~\ref{sec:inductive_bias} confirm that both variants exhibit strong spatial inductive bias in practice.

\subsection{Preliminaries}
\label{sec:preliminaries}

State Space Models (SSMs) map an input sequence $x(t) \in \mathbb{R}$ to an output $y(t) \in \mathbb{R}$ through a latent state $h(t) \in \mathbb{R}^N$ , according to the following dynamics:%
\vspace{-5pt}
\begin{align}\nonumber
    h'(t) = \mathbf{A}h(t) + \mathbf{B}x(t), \quad %\\
    y(t) = \mathbf{C}h(t) + \mathbf{D}x(t),
\end{align}
where $\mathbf{A} \in \mathbb{R}^{N \times N}$, $\mathbf{B} \in \mathbb{R}^{N \times 1}$, $\mathbf{C} \in \mathbb{R}^{1 \times N}$, and $\mathbf{D} \in \mathbb{R}$ are learnable parameters. For discrete sequences, the system is discretized using step size $\Delta$:%
\vspace{-5pt}
\begin{align}\nonumber
    h_t = \bar{\mathbf{A}}h_{t-1} + \bar{\mathbf{B}}x_t, \quad%\\
    y_t = \mathbf{C}h_t + \mathbf{D}x_t,
\end{align}
where $\bar{\mathbf{A}}$ and $\bar{\mathbf{B}}$ are the discretized parameters.
Intuitively, $h_t$ acts as a memory summarizing the input history, where $\bar{\mathbf{A}}$ sets the forgetting rate and $\bar{\mathbf{B}}$/$\mathbf{C}$ control what is written to and read from it.
Mamba~\cite{gu2024mamba} introduces input-dependent selection by making $\mathbf{B}$, $\mathbf{C}$, and $\Delta$ functions of input $x_t$, enabling content-aware reasoning with linear complexity. Mamba-2~\cite{dao2024mamba2} further constrains $\mathbf{A}$ to a scalar times identity matrix, recasting computation as structured matrix multiplications to gain 2-8$\times$ speedup.

\subsection{Model Architecture}
\label{sec:architecture}

As shown in Figure~\ref{fig:block_diagram}, \method follows CLIP's dual-encoder architecture, mapping images and text into a shared embedding space. Both encoders are Mamba-based, offering consistent sub-quadratic memory scaling across modalities.

\noindent \textbf{Vision Encoder.}
We use VMamba~\cite{liu2024vmamba} as the vision encoder. Given an input image $\mathbf{I} \in \mathbb{R}^{H \times W \times 3}$, we divide it into non-overlapping patches of size $P \times P$ and project them into patch embeddings processed through Visual State-Space (VSS) blocks utilizing the SS2D cross-scan mechanism. The hierarchical structure progressively downsamples feature maps through patch merging, with final features mapped to the shared embedding space via a learned projection $W_v$.

A key advantage is the implicit handling of spatial relationships through scanning patterns, providing a favorable spatial inductive bias (see Section~\ref{sec:inductive_bias}), allowing \method to process variable input resolutions without positional encoding interpolation, specialized schemes like RoPE~\cite{su2024roformer}, or complex training procedures like FlexViT~\cite{beyer2023flexvit} and NaFlex~\cite{dehghani2023patch}.

\noindent \textbf{Text Encoder.}
We employ two pretrained variants: Mamba-1~\cite{gu2024mamba} (1.4B parameters) and Mamba-2~\cite{dao2024mamba2} (1.3B parameters). While both share the selective state-space foundation, Mamba-2 introduces structured state-space duality (SSD) enabling more efficient computation. Unlike Transformers with mature bidirectional encoders~\cite{bert, liu2019roberta}, Mamba models are autoregressive by design. However, this autoregressive formulation is well-suited for contrastive learning: the last-token representation naturally aggregates full-context information, and the absence of a fixed positional encoding enables processing of arbitrarily long text - a key advantage for dense captioning retrieval (Section~\ref{sec:dense_captioning_retrieval}). We further analyze the role of the text tower in Table~\ref{tab:llm_ablation}.

Given tokenized input $\mathbf{T} = [t_1, t_2, \ldots, t_L]$ with padding mask $\mathbf{m} \in \{0,1\}^L$, we extract the hidden state at the last non-padding token as the text representation:
\begin{equation}
\mathbf{t}_{\text{raw}} = \mathbf{H}_{k}, \quad \text{where}\ \, k = \max\{i : m_i = 0\}
\end{equation}
where $\mathbf{H} = [\mathbf{h}_1, \mathbf{h}_2, \ldots, \mathbf{h}_L]$ denotes hidden states from the last Mamba layer.
This last-token pooling is well-suited for Mamba's causal formulation, where each $\mathbf{h}_t$ is computed recurrently, making the last token the only position with full context access. The representation is projected to the shared embedding space via $W_t$.

%% file: tables/main_results.tex
\begin{table*}[t]
\centering
\small
\resizebox{0.8\textwidth}{!}{%
\begin{tabular}{llccccc}
\toprule
& & \multicolumn{2}{c}{Zero-shot Classification} & \multicolumn{2}{c}{Zero-shot Retrieval} \\
\cmidrule(lr){3-4} \cmidrule(lr){5-6}
Vision Tower & Text Tower & Acc@1 & Acc@5 & IR@5 & TR@5 \\
\midrule
FlexViT-B/16~\cite{beyer2023flexvit} & LLaMA-3.2~\cite{touvron2023llama} & 26.3 & 57.4 & 62.4 & 72.3 \\
NaFlex-B/16~\cite{dehghani2023patch} & LLaMA-3.2~\cite{touvron2023llama} & 26.1 & 56.6 & 61.5 & 73.3 \\
ViT-B/16~\cite{dosovitskiy2021image} & LLaMA-3.2~\cite{touvron2023llama} & \underline{27.3} & 56.4 & 62.8 & 72.9 \\
RoPE-ViT-B/16~\cite{heo2024ropevit} & LLaMA-3.2~\cite{touvron2023llama} & \underline{27.3} & \underline{59.0} & 63.4 & 72.9 \\
\hdashline
MetaCLIP-ViT-B/16$^\dagger$~\cite{xu2024demystifying} & LLaMA-3.2~\cite{touvron2023llama} & 25.7 & 56.9 & \textbf{66.4} & \underline{76.2} \\
\midrule
\method (VMamba-B~\cite{liu2024vmamba}) & Mamba-2~\cite{dao2024mamba2} & 26.9 & \textbf{59.0} & {65.2} & {75.3} \\
\method (VMamba-B)~\cite{liu2024vmamba}) & Mamba-1~\cite{gu2024mamba} & \textbf{29.6} & 58.5 & \underline{65.5} & \textbf{77.0} \\
\bottomrule
\end{tabular}%
}
\caption{\textbf{CLIP-Benchmark Results.} Average Results for Zero-shot classification and retrieval performance on 31 datasets (28 for classification, 3 for retrieval; full details in the supplementary material). Both \method variants achieve the top retrieval performance among models with comparable pretraining, outperforming all transformer baselines. For classification, the Mamba-1 variant achieves the best Acc@1, while the Mamba-2 variant ties for the best Acc@5. $^\dagger$MetaCLIP-ViT uses a vision encoder pretrained on MetaCLIP-400M~\cite{xu2024demystifying} (400$\times$ larger than IN-1K); despite this advantage, it achieves lower classification accuracy than \method while reaching comparable retrieval.}
\label{tab:main_results}
\vspace{-16pt}
\end{table*}

%% file: results.tex
\input{tables/ood_robustness}
\section{Results}
\label{sec:results}
We evaluate \method across standard CLIP benchmarks (Section~\ref{subsec::clip_benchmarks}), out-of-distribution robustness (Section~\ref{subsec::ood_robustness}), resolution flexibility (Section~\ref{subsec::high_resolution}), dense captioning retrieval (Section~\ref{sec:dense_captioning_retrieval}), and provide analysis of representational geometry, efficiency, scaling, and spatial inductive bias (Section~\ref{subsec::analysis}).

\noindent \textbf{Experimental Setup.}
We train all models on CC12M~\cite{cc12m} for 10 epochs at $224 \times 224$ resolution using AdamW with cosine learning rate schedule (peak LR $5 \times 10^{-5}$), batch size 2048, and projection dimension 768. All vision encoders are base-sized ($\sim$86M parameters) initialized from ImageNet-1K~\cite{deng2009imagenet}. For \method, we use VMamba-B~\cite{liu2024vmamba} as vision encoder with two text encoder variants: Mamba-1 (1.4B)~\cite{gu2024mamba} and Mamba-2 (1.3B)~\cite{dao2024mamba2}. We compare against three transformer baselines, all using LLaMA-3.2-1B~\cite{touvron2023llama} as text encoder: RoPE-ViT~\cite{heo2024ropevit}, FlexViT~\cite{beyer2023flexvit}, and NaFlex-ViT~\cite{dehghani2023patch}. To further strengthen the comparison, we include a MetaCLIP-ViT-B/16 baseline pretrained on MetaCLIP-400M~\cite{xu2024demystifying} - a dataset 400$\times$ larger than ImageNet-1K.

Our evaluation follows the controlled-comparison paradigm adopted in recent vision-language research~\cite{liu-etal-2025-data-language}: all models are trained on the same dataset with identical protocols, isolating architectural differences from confounding factors such as dataset scale or curation. Our scaling experiments (Section~\ref{sec:scaling}) confirm that \method's advantages are consistent across data and model sizes with unsaturated performance curves, indicating the scalability of our results.

\subsection{CLIP-Benchmarks Results}
\label{subsec::clip_benchmarks}

We evaluate on CLIP-Benchmark~\cite{clip_benchmarks} (31 datasets for zero-shot classification and retrieval). Table~\ref{tab:main_results} presents our main findings. Both \method variants achieve top retrieval performance: Mamba-1 leads with 65.5\% image and 77.0\% text recall, outperforming the best transformer baseline (RoPE-ViT) by +2.1\% and +4.1\% respectively. For classification, \method-Mamba-1 achieves the best Acc@1 (29.6\%, +2.3\% over the best transformer), while \method-Mamba-2 ties for best Acc@5 (59.0\%). Notably, MetaCLIP-ViT - despite vision pretraining on 400$\times$ more data - achieves lower classification accuracy (25.7\%) and comparable retrieval, suggesting that \method's advantages stem from architectural properties rather than data scale.
We further stress the architectural gains by comparing \method{} to convolution based variants (i.e CLIP based ResNets~\cite{He2015DeepRL}), We evaluate on Open-CLIP~\cite{open_clip} 38 benchmarks, \method{} outperformed all ResNet based models more than 4 points on average on both zero-shot classification and retrieval, outperforming the strongest ResNet variant. Full details are in the supplementary materials.

Between our two variants, Mamba-1 consistently outperforms Mamba-2 on most metrics. While Mamba-2 achieves 2--8$\times$ computational speedups through key simplifications - restricting the transition matrix from a structured diagonal (HiPPO-initialized) to a scalar identity, and adopting multi-head weight sharing - these reduce per-token expressivity. Mamba-1's per-channel selective scan, with fully input-dependent parameters ($\Delta$, $\mathbf{B}$, $\mathbf{C}$), provides finer-grained filtering that has more discriminative representations for contrastive alignment.

\subsection{OOD Robustness}
\label{subsec::ood_robustness}

Having established \method's retrieval advantages, we next examine whether its representations also improve robustness under distribution shift. We evaluate on five ImageNet variants: ImageNet-V2~\cite{imagenetv2}, ImageNet-R~\cite{imagenet_r}, ImageNet-A~\cite{imagenet_ao}, ImageNet-O~\cite{imagenet_ao}, and ImageNet-Sketch~\cite{imagenet_sketch}.

Table~\ref{tab:ood_robustness} shows that \method variants achieve the top two average robustness scores, with Mamba-1 leading at 35.2\% top-1 and 64.3\% top-5, outperforming the best transformer baseline (RoPE-ViT) by +2.0\% and +2.2\% respectively.

\input{tables/resolution}

The most striking result is on ImageNet-O. Both \method variants dramatically outperform all transformer baselines. Mamba-2 achieves 49.8\% and Mamba-1 48.1\% top-1 accuracy - surpassing the best transformer by +9.7\% and +8.0\%. Remarkably, both variants also surpass CLIP-ViT-B-16~\cite{radford2021learning} (42.3\%)\footnote{Taken from: \href{https://github.com/mlfoundations/open_clip/blob/main/docs/openclip_results.csv}{OpenCLIP Results} (March 2026)} trained on LAION-2B~\cite{schuhmann2022laion}, a dataset 167$\times$ larger than CC12M. This provides direct evidence that architectural inductive biases can outweigh data scale for OOD robustness, consistent with our geometric analysis (Section~\ref{sec:geometry}) showing that \method's reduced hubness leads to less reliance on spurious feature correlations.

Both \method variants also lead on ImageNet-V2 (+3.1\%/+2.6\%) and Sketch. On ImageNet-R and ImageNet-A, RoPE-ViT performs best, though \method remains competitive within 1--2\%, consistent with prior findings that SSMs and transformers exhibit complementary failure modes~\cite{du2024understanding}.

\input{tables/appendix_high_res_ret}

\subsection{High-Resolution}
\label{subsec::high_resolution}

We next evaluate resolution flexibility - a practical requirement for applications that process images at their native resolution. Unlike transformers that require positional embedding interpolation (RoPE-ViT) or specialized training (FlexViT, NaFlex), \method generalizes to higher resolutions without architectural modifications or additional training. We train all models at 224$\times$224 and evaluate at up to 896$\times$896 with no fine-tuning.

\noindent \textbf{CLIP-Benchmark Evaluation.} We evaluate on CLIP-Benchmark at resolutions up to 384$\times$384, matching the native resolution range of its constituent datasets.
Table~\ref{tab:resolution_comparison} shows that both \method variants achieve the top retrieval performance across all resolutions, with Mamba-1 achieving the best Acc@1 throughout. This advantage persists at 320$\times$320 and 384$\times$384 without any resolution-specific adaptation.

\input{tables/appendix_flickr8k_rephrased}
\input{tables/appendix_docci}

\noindent \textbf{Scaling to Higher Resolutions.} To rigorously evaluate resolution scaling beyond standard benchmarks, we test on NoCaps~\cite{agrawal2019nocaps} (4.5K images, avg.\ 810$\times$960) and Crossmodal-3600~\cite{ThapliyalCrossmodal2022} (3.6K images, avg.\ 640$\times$520) - benchmarks whose native resolutions substantially exceed 224$\times$224. While performance decreases at higher resolutions for all models - as expected when evaluating at 2--4$\times$ the training resolution - \method degrades far more gracefully.
Table~\ref{tab:appendix_high_res_retrieval_results} shows retrieval performance across resolutions. Both \method variants outperform all baselines at every resolution, with the gap widening substantially at higher resolutions. At 896$\times$896, \method achieves +18--19\% advantage over RoPE-ViT on both image and text retrieval. Notably, both variants surpass FlexViT and NaFlex despite these models being specifically designed for variable resolutions.

\subsection{Dense Captioning Retrieval}
\label{sec:dense_captioning_retrieval}

% Complementing our high-resolution evaluation, we evaluate robustness to out-of-distribution text lengths. \method overcomes CLIP's fixed 77-token context window through Mamba's autoregressive text encoder, which naturally handles arbitrarily long sequences. We evaluate on: (1) Flickr8k-test captions rephrased using LLaMA-3.3-70B~\cite{touvron2023llama} (average 134 tokens, 98.3\% exceeding 77 tokens), with images at 224$\times$224, and (2) DOCCI~\cite{OnoeDocci2024} with naturally verbose descriptions (average 142 tokens, 94.4\% exceeding 77 tokens), with images at 896$\times$896. As shown in Tables~\ref{tab:flickr8k_rephrased} and~\ref{tab:docci}, \method-Mamba-1 achieves the best image retrieval across both benchmarks, with up to 6.3\% improvement on Flickr8k-Rephrased and 8.2\% on DOCCI over the best transformer baseline. On DOCCI text retrieval, NaFlex achieves the highest scores, likely benefiting from its sequence packing design; \method remains competitive on this metric while substantially leading on image retrieval.

Complementing our high-resolution evaluation, we evaluate robustness to out-of-distribution text lengths. \method overcomes the standard CLIP's fixed 77-token context window through Mamba's autoregressive text encoder, which naturally handles arbitrarily long sequences. 
Note that in our evaluation, no method is restricted to 77 tokens at evaluation - all transformer baselines use LLaMA-3.2-1B (8K native context) and Mamba LLMs support arbitrary length, so long captions are processed in full by every model.
We evaluate on: (1) Flickr8k-test captions rephrased using LLaMA-3.3-70B~\cite{touvron2023llama} (average 134 tokens, 98.3\% exceeding 77 tokens), with images at 224$\times$224, and (2) DOCCI~\cite{OnoeDocci2024} with naturally verbose descriptions (average 142 tokens, 94.4\% exceeding 77 tokens), with images at 896$\times$896. As shown in Tables~\ref{tab:flickr8k_rephrased} and~\ref{tab:docci}, \method-Mamba-1 achieves the best image retrieval across both benchmarks, with up to 6.3\% improvement on Flickr8k-Rephrased and 8.2\% on DOCCI over the best transformer baseline. On DOCCI text retrieval, NaFlex achieves the highest scores, likely benefiting from its sequence packing design; \method remains competitive on this metric while substantially leading on image retrieval.

\input{tables/inductive_bias}
\subsection{Analysis}
\label{subsec::analysis}
\vspace{-1pt}
\subsubsection{Spatial Inductive Bias}
\label{sec:inductive_bias}
To understand the source of VMamba's advantages, we investigate its spatial inductive bias. Prior work has shown that locality bias benefits vision tasks by enabling more sample-efficient learning \cite{dascoli2021convit}, a property ViTs possess only weakly through their positional encoding (PE). We validate this using lightweight 3-layer VMamba (0.33M params) and ViT (0.35M params) models trained on CIFAR-10 \cite{krizhevsky2009cifar} with regular versus spatially distorted data with shuffled patch orders. VMamba achieves lower training loss on regular images, while ViT performs better on the distorted variant, confirming that SSM-based encoders encode sequential spatial structure as a much stronger inductive bias. Full results appear in Table~\ref{tab:inductive_bias}.

% To understand the source of VMamba's advantages, we investigate its spatial inductive bias. Prior work has shown that locality bias benefits vision tasks by enabling more sample-efficient learning \cite{dascoli2021convit}, a property ViTs lack due to their permutation-invariant attention. We validate this using lightweight 3-layer VMamba (0.33M params) and ViT (0.35M params) models trained on CIFAR-10 \cite{krizhevsky2009cifar} with regular versus spatially distorted data with shuffled patch orders. VMamba achieves lower training loss on regular images, while ViT performs better on the distorted variant, confirming that SSM-based encoders encode sequential spatial structure as an inductive bias. Full results appear in Table~\ref{tab:inductive_bias}.

\subsubsection{Qualitative Results}
\method produces more interpretable image-text alignment maps. As shown in Figure~\ref{fig:nocaps_visuals}, \method correctly localizes the wooden deck and fence structure described in the caption, while RoPE-ViT and FlexViT exhibit diffuse attention scattered across irrelevant areas.
\input{figures/nocaps_visuals}

\subsubsection{Representational Geometry Analysis}
\label{sec:geometry}
To understand \method's retrieval advantages, we analyze embedding geometry on NoCaps~\cite{agrawal2019nocaps}. Following \cite{wang2020understanding}, we measure \textit{alignment} (matched pairs should be close), \textit{uniformity} (embeddings spread evenly on the hypersphere), and \textit{hubness}~\cite{radovanovic2010hubs} - the tendency for certain embeddings to dominate nearest-neighbor lists, degrading retrieval quality~\cite{zhang2024adversarial}.

\input{tables/geometric_retrieval}

Table~\ref{tab:geometry} shows that \method achieves the best alignment (0.982) and lowest text hubness (1.13), directly explaining its retrieval gains: better alignment improves matching accuracy, while reduced text hubness benefits text-to-image retrieval. On image-side metrics, NaFlex achieves the best image uniformity and hubness, suggesting complementary strengths across architectures. The lower text hubness also implies less reliance on generic, spuriously correlated features, which may contribute to \method's improved OOD robustness (Section~\ref{subsec::ood_robustness}).

% \subsubsection{\method{} vs. Tansformer-CLIP: Qualitative Insights}
% A fair comparison requires articulating not only where CLIMP wins, but where transformers retain an edge and where CLIMP fails. Using the CLIP-Benchmark suite, we organize both sides by the architectures' internal mechanisms and prior analyses of visual SSMs, giving a basis for choosing between them.

% \noindent\textbf{CLIMP outperforms transformers in:} 
% (1) \textit{High resolution.} The largest and most consistent advantage. The gap to transformer baselines widens with input resolution; by 896$\times$896 CLIMP exceeds RoPE-ViT on retrieval, driven by Mamba's native variable-resolution support. 
% (2) \textit{Sparse, distributed perturbations.} CLIMP's scanning-based receptive field dilutes distributed spurious cues, while attention can latch onto them locally; consistent with the ImageNet-O gain, see~\ref{tab:ood_robustness}.

% \noindent\textbf{Transformers may be favorable in:} 
% (1) \textit{Low-resolution / small-image inputs.} Few patches mean fewer scanning steps for context aggregation, while ViT's global attention is comparatively unaffected.  
% (2) \textit{Strong visual-style shift.} VMamba's spatial features may transfer less well than ViT's more abstract attention features, consistent with RoPE-ViT's slight lead on ImageNet-R.
% Visual examples can be found in the supplementary materials.

\subsubsection{\method{} vs. Transformer-CLIP: Qualitative Insights}
A fair comparison requires articulating not only where CLIMP wins, but where transformers retain an edge. Using the CLIP-Benchmark suite, we organize both sides by the architectures' internal mechanisms and prior analyses of visual SSMs.
\noindent\textbf{CLIMP outperforms transformers in:} 
(1) \textit{High resolution.} The largest and most consistent advantage: the gap widens with input resolution, and by 896$\times$896 CLIMP exceeds RoPE-ViT on retrieval, driven by Mamba's native variable-resolution support. 
(2) \textit{Sparse, distributed perturbations.} CLIMP's scanning-based receptive field dilutes distributed spurious cues, while attention latches onto them locally; consistent with the ImageNet-O gain, see~\ref{tab:ood_robustness}.
\noindent\textbf{Transformers may be favorable in:} 
(1) \textit{Low-resolution inputs.} Few patches mean fewer scanning steps for context aggregation, while ViT's global attention is comparatively unaffected.  
(2) \textit{Strong visual-style shift.} VMamba's spatial features may transfer less well than ViT's abstract attention features, consistent with RoPE-ViT's slight lead on ImageNet-R.
Visual examples appear in the supplementary materials.
\subsubsection{Memory and Computational Efficiency}
\input{figures/memory}
\method offers significant efficiency advantages at high resolutions. As shown in Figure~\ref{fig:memory_flops}, \method requires 5$\times$ less resolution-specific memory (10.0 vs 50.4 MB) and 1.8$\times$ fewer FLOPs (259.7 vs 457.9 GFLOPs) than ViT variants at 896$\times$896, with the gap widening at higher resolutions due to Mamba's linear complexity.

\subsubsection{Scaling}
\label{sec:scaling}
We investigate how \method scales with model size and training data, directly addressing whether its advantages persist at larger scales.

\noindent \emph{Model Size.} Table~\ref{tab:model_size} shows ImageNet-1K zero-shot classification across model scales (22M--87M parameters). \method consistently outperforms ViT-based alternatives at every scale, with the advantage maintained from the smallest (22M: +5.6\% over FlexViT) to the largest (87M: +2.8\% over RoPE-ViT) configuration.
\input{tables/model_size}

\noindent \emph{Dataset Size.} Figure~\ref{fig:scaling_laws} examines scaling with training data on Conceptual Captions. \method exhibits a steep scaling curve that has not saturated at 12M samples, suggesting continued benefits with larger datasets. The consistent performance gap over transformer baselines across all data sizes (1M, 3M, 12M) indicates that \method's advantages stem from architectural properties rather than overfitting to a particular data regime. This is further supported by the MetaCLIP-ViT comparison (Table~\ref{tab:main_results}): despite vision pretraining on 400$\times$ more data, it achieves lower classification accuracy than \method, providing direct evidence that our architectural advantages extend beyond the training scale.
\input{figures/scaling_laws}

\vspace{-2pt}
\subsection{Ablations}
\label{sec:ablations}
\input{tables/ablation_llm}
We ablate the text encoder contribution by replacing the LLM backbone while keeping the vision encoder fixed. As shown in Table~\ref{tab:llm_ablation}, the results reveal an architectural synergy: while RoPE-ViT shows minimal sensitivity to text encoder choice ($\pm$1\% across metrics), VMamba consistently benefits from Mamba-based text encoders, achieving +3.5\% Acc@1 and +3.9\% TR@5 over LLaMA. This suggests that matched SSM architectures in both modalities learn more compatible representations for cross-modal alignment. We additionally evaluate Hydra~\cite{hwang2024hydra}, a bidirectional SSM variant, which underperforms the autoregressive Mamba LLMs - confirming that mature pretrained language models provide stronger text representations than current bidirectional SSM alternatives.

%% file: tables/ood_robustness.tex
\begin{table*}[!t]
\centering
\resizebox{\textwidth}{!}{%
\begin{tabular}{llcccccc}
\toprule
Vision Tower & Text Tower & IN-V2 & IN-R & IN-A & IN-O & Sketch & Avg \\
\midrule
FlexViT-B/16 & LLaMA-3.2 & 32.8/62.5 & 45.4/74.3 & 13.2/41.2 & 38.1/68.2 & 24.6/50.0 & 30.8/59.2 \\
NaFlex-B/16 & LLaMA-3.2 & 31.4/60.5 & 44.8/72.5 & 12.0/36.0 & 38.5/67.6 & 23.5/49.1 & 30.0/57.1 \\
ViT-B/16 & LLaMA-3.2 & 30.0/57.5 & 42.9/69.7 & 13.9/37.6 & 27.7/55.8 & 21.7/46.1 & 27.2/53.3 \\
RoPE-ViT-B/16 & LLaMA-3.2 & 34.4/65.4 & \textbf{47.8}/\textbf{76.0} & \textbf{16.3}/\textbf{46.3} & 40.1/69.9 & \underline{27.4}/53.1 & 33.2/62.1 \\
\midrule
\method (VMamba-B) & Mamba-2 & \underline{37.0}/\underline{67.6} & \underline{46.6}/\underline{74.5} & \underline{15.6}/\underline{45.5} & \textbf{49.8}/\underline{77.0} & 27.0/\underline{54.2} & \underline{34.8}/\underline{63.7} \\
\method (VMamba-B) & Mamba-1 & \textbf{37.5}/\textbf{68.4} & 46.2/74.5 & 15.5/45.3 & \underline{48.1}/\textbf{78.8} & \textbf{27.5}/\textbf{54.4} & \textbf{35.2}/\textbf{64.3} \\
\bottomrule
\end{tabular}%
}
\caption{\textbf{Out-of-distribution robustness.} Evaluation on ImageNet variant datasets, reporting top-1/top-5 accuracy. Both \method variants achieve the top two average scores, with particularly strong gains on ImageNet-O (+9.7\%/+8.0\% over the best transformer) and ImageNet-V2 (+3.1\%/+2.6\%).}
\label{tab:ood_robustness}
\vspace{-16pt}
\end{table*} 

%% file: tables/resolution.tex
\begin{table}[t]
\centering
\setlength{\tabcolsep}{4pt}
\resizebox{\columnwidth}{!}{%
\begin{tabular}{ll ccc ccc}
\toprule
& & \multicolumn{3}{c}{\textbf{Acc@1/Acc@5}} & \multicolumn{3}{c}{\textbf{IR@5/TR@5}} \\
\cmidrule(lr){3-5} \cmidrule(lr){6-8}
\textbf{Vision} & \textbf{Text} & 224 & 320 & 384 & 224 & 320 & 384 \\
\midrule
FlexViT  & LLaMA   & 26.3/57.3 & 26.1/57.3 & 25.9/\underline{56.9} & 62.4/72.3 & 63.5/74.3 & 63.6/\underline{75.0} \\
NaFlex   & LLaMA   & 26.1/56.6 & 25.5/56.6 & 25.1/56.1 & 61.4/73.2 & 63.0/74.1 & 63.4/74.6 \\
RoPE-ViT & LLaMA   & 27.3/\textbf{59.0} & 26.4/\textbf{58.4} & 24.9/\textbf{57.1} & 63.3/72.9 & 64.1/74.3 & 63.3/73.2 \\
\midrule
VMamba   & Mamba-2 & 26.9/\textbf{59.0} & 26.1/\underline{58.2} & 24.9/\underline{56.9} & \underline{65.2}/\underline{75.2} & \textbf{66.2}/\underline{76.2} & \textbf{65.2}/74.6 \\
VMamba   & Mamba-1 & \textbf{29.6}/58.5 & \textbf{28.4}/57.5 & \textbf{27.3}/56.3 & \textbf{65.5}/\textbf{77.0} & \underline{66.0}/\textbf{76.9} & \underline{65.0}/\textbf{75.2} \\
\bottomrule
\end{tabular}
}
\caption{\textbf{Resolution scaling on CLIP-Benchmark.} All models support arbitrary resolution inference. Both \method variants demonstrate strong performance across all resolutions, achieving the best results on Acc@1 and retrieval metrics (IR@5/TR@5).}
\label{tab:resolution_comparison}
\vspace{-16pt}
\end{table}

%% file: tables/appendix_high_res_ret.tex
\begin{table}[t]
\centering
\setlength{\tabcolsep}{3pt}
\resizebox{\columnwidth}{!}{%
\begin{tabular}{ll|cc|cc|cc|cc|cc|cc}
\toprule
& & \multicolumn{6}{c|}{\textbf{NoCaps}} & \multicolumn{6}{c}{\textbf{Crossmodal-3600}} \\
& & \multicolumn{2}{c|}{224} & \multicolumn{2}{c|}{512} & \multicolumn{2}{c|}{896} & \multicolumn{2}{c|}{224} & \multicolumn{2}{c|}{512} & \multicolumn{2}{c}{896} \\
\textbf{Vision} & \textbf{Text} & I$\rightarrow$T & T$\rightarrow$I & I$\rightarrow$T & T$\rightarrow$I & I$\rightarrow$T & T$\rightarrow$I & I$\rightarrow$T & T$\rightarrow$I & I$\rightarrow$T & T$\rightarrow$I & I$\rightarrow$T & T$\rightarrow$I \\
\midrule
RoPE-ViT & LLaMA   & 67.6 & 78.9 & 65.1 & \textbf{75.9} & 38.6 & 34.9 & 64.4 & 68.2 & 62.3 & \textbf{64.1} & 36.8 & 26.9 \\
NaFlex   & LLaMA   & 60.4 & 69.1 & 59.7 & 65.0 & 51.6 & 50.8 & 56.7 & 59.3 & 56.3 & 55.7 & 48.2 & 42.0 \\
FlexViT  & LLaMA   & 56.0 & 62.9 & 53.4 & 57.8 & 42.2 & 41.2 & 52.6 & 50.5 & 50.9 & 47.6 & 40.4 & 34.1 \\
\midrule
CLIMP(VMamba)   & Mamba-2 & \underline{70.2} & \underline{81.3} & \underline{66.4} & 72.4 & \textbf{57.1} & \textbf{53.8} & \textbf{65.2} & \textbf{69.2} & \textbf{62.8} & \underline{61.9} & \underline{54.7} & \textbf{46.0} \\
CLIMP(VMamba)   & Mamba-1 & \textbf{70.5} & \textbf{81.9} & \textbf{66.9} & \underline{72.7} & \underline{57.1} & \underline{53.5} & \underline{65.1} & \underline{69.1} & \underline{62.6} & 60.8 & \textbf{55.9} & \underline{45.4} \\
\bottomrule
\end{tabular}%
}
\caption{Retrieval Recall@5 (IR@5/TR@5) on NoCaps~\cite{agrawal2019nocaps} and Crossmodal-3600~\cite{ThapliyalCrossmodal2022} at different resolutions for both image to text (I$\rightarrow$T) and text to image (T$\rightarrow$I). Both \method variants demonstrate strong performance across all resolutions, with the performance gap over transformer baselines widening substantially at 896$\times$896.}
\label{tab:appendix_high_res_retrieval_results}
\vspace{-10pt}
\end{table}

%% file: tables/appendix_flickr8k_rephrased.tex
\begin{table}[t]
\centering
\begin{tabular}{ll cccccc}
\toprule
& & \multicolumn{3}{c}{\textbf{Image Retrieval}} & \multicolumn{3}{c}{\textbf{Text Retrieval}} \\
\cmidrule(lr){3-5} \cmidrule(lr){6-8}
\textbf{Vision Tower} & \textbf{Text Tower} & R@1 & R@5 & R@10 & R@1 & R@5 & R@10 \\
\midrule
FlexViT  & LLaMA & 46.4 & 75.0 & 84.5 & 55.7 & 79.0 & 87.8 \\
NaFlex   & LLaMA & 50.9 & 80.2 & 88.9 & 66.1 & 86.8 & 93.8 \\
ViT      & LLaMA & 53.9 & 82.0 & 90.1 & 64.8 & 85.8 & 92.0 \\
RoPE-ViT & LLaMA & 60.7 & 85.8 & 93.3 & 77.6 & 93.6 & 97.0 \\
\midrule
CLIMP (VMamba)   & Mamba-2 & 63.3 & 88.2 & 94.6 & 75.7 & 92.6 & 96.2 \\
CLIMP (VMamba)   & Mamba-1 & \textbf{67.0} & \textbf{89.4} & \textbf{94.7} & \textbf{81.3} & \textbf{95.9} & \textbf{98.4} \\
\bottomrule
\end{tabular}%
\caption{\textbf{Zero-shot retrieval on rephrased Flickr8k-test with dense captions}. To make the captions dense, the original captions were rephrased using an LLM (avg.\ 134 tokens vs.\ original 14), with 98.32\% of them exceeding the 77-token limit. Our \method models consistently outperform transformer-based baselines, demonstrating effective retrieval with extended textual descriptions beyond CLIP's 77-token limit.}
\label{tab:flickr8k_rephrased}
\vspace{-10pt}
\end{table}

%% file: tables/appendix_docci.tex
\begin{table}[t]
\centering
\begin{tabular}{ll cccccc}
\toprule
& & \multicolumn{3}{c}{\textbf{Image Retrieval}} & \multicolumn{3}{c}{\textbf{Text Retrieval}} \\
\cmidrule(lr){3-5} \cmidrule(lr){6-8}
\textbf{Vision Tower} & \textbf{Text Tower} & R@1 & R@5 & R@10 & R@1 & R@5 & R@10 \\
\midrule
RoPE-ViT & LLaMA   & 18.0 & 39.5 & 50.5 & 10.5 & 27.9 & 38.1 \\
FlexViT  & LLaMA   & 20.5 & 42.8 & 53.8 & 15.6 & 36.3 & 47.6 \\
NaFlex   & LLaMA   & 29.1 & 55.3 & 67.5 & \textbf{26.1} & \textbf{53.4} & \textbf{65.1} \\
\midrule
CLIMP (VMamba)   & Mamba-2 & 33.4 & 62.6 & 74.0 & 24.6 & 51.7 & 63.7 \\
CLIMP (VMamba)   & Mamba-1 & \textbf{37.3} & \textbf{67.1} & \textbf{77.8} & 23.8 & 50.4 & 62.7 \\
\bottomrule
\end{tabular}%
\caption{\textbf{Zero-shot retrieval on DOCCI~\cite{OnoeDocci2024}.} DOCCI is a dense captioning dataset with long descriptions (avg.\ 142 tokens, 94.4\% exceeding 77 tokens). Our \method models achieve the best image retrieval across all metrics. On text retrieval, \method remains competitive but NaFlex leads, likely benefiting from its sequence packing design.}
\label{tab:docci}
\vspace{-10pt}
\end{table}

%% file: tables/inductive_bias.tex
\begin{table}[t]
\centering
\begin{tabular}{llcc}
\toprule
\textbf{Model} & \textbf{Data} & \textbf{Train Loss} & \textbf{Test Acc.}\\
\midrule
VMamba & Regular  & \textbf{1.028} & \textbf{69.33} \\
ViT    & Regular  & 1.121 & 67.50\\
\midrule
VMamba & Shuffled & 1.612 & 46.44 \\
ViT    & Shuffled & \textbf{1.328} & \textbf{59.72} \\
\bottomrule
\end{tabular}
\caption{Spatial inductive bias analysis. We train lightweight 3-layer VMamba (0.33M params) and ViT (0.35M params) on CIFAR-10 with regular and shuffled patch orders. VMamba achieves lower training loss on regular images but degrades significantly under shuffling, while ViT shows the opposite pattern—performing relatively better on distorted data. This confirms that SSM-based encoders rely on sequential spatial structure as an inductive bias.}
\label{tab:inductive_bias}
\vspace{-10pt}
\end{table}

%% file: figures/nocaps_visuals.tex
\begin{figure*}[t]
    \centering
    \setlength{\tabcolsep}{1pt}
    \begin{tabular}{cccc}             
        Model (similarity) & Flex-ViT (0.452) & RoPE-ViT (0.501) & \method (0.545) \\
        \includegraphics[width=0.24\textwidth]{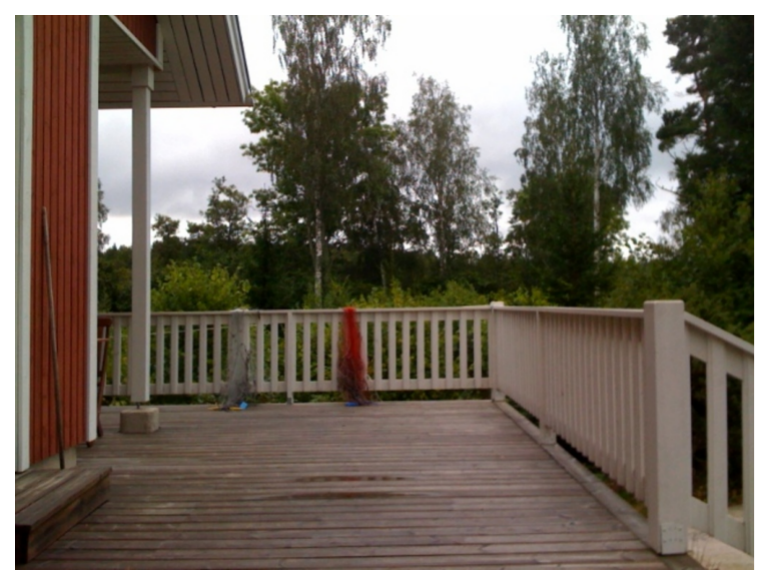} &
        \includegraphics[width=0.24\textwidth]{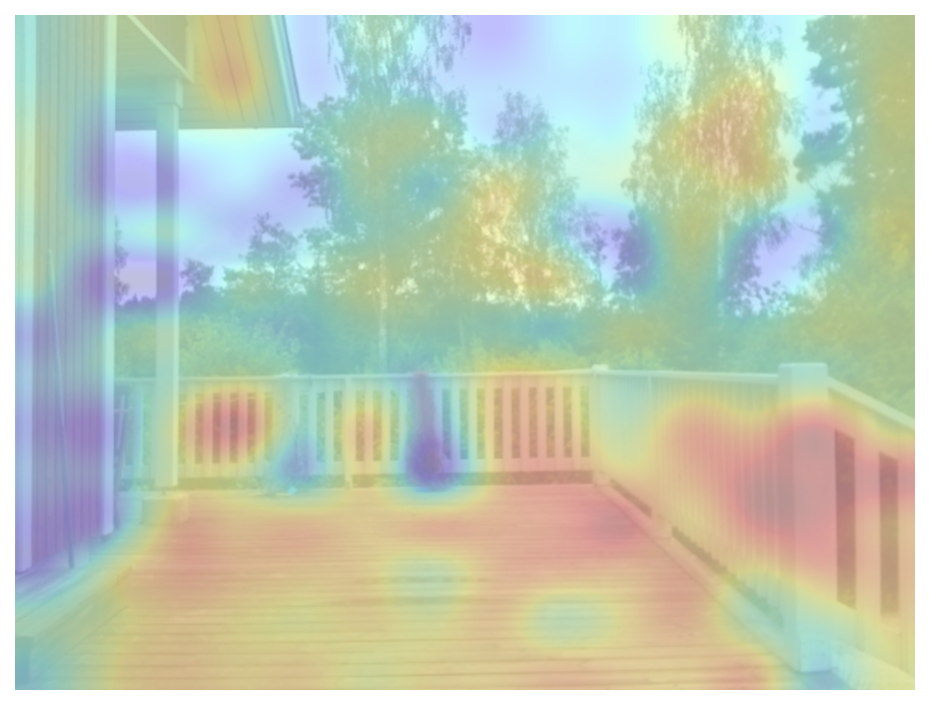} &
        \includegraphics[width=0.24\textwidth]{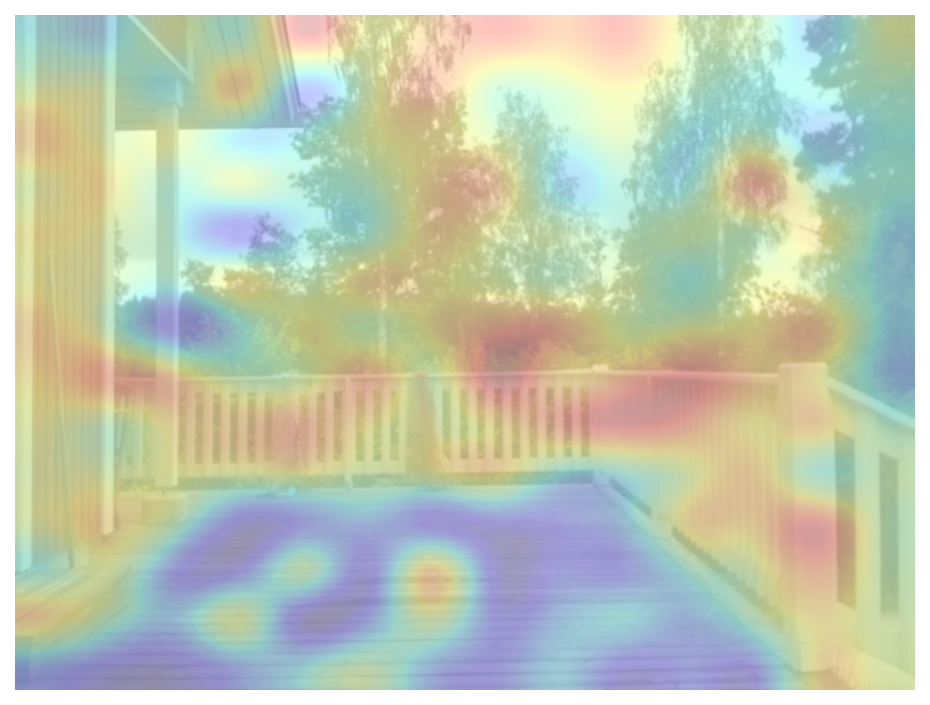} &
        \includegraphics[width=0.24\textwidth]{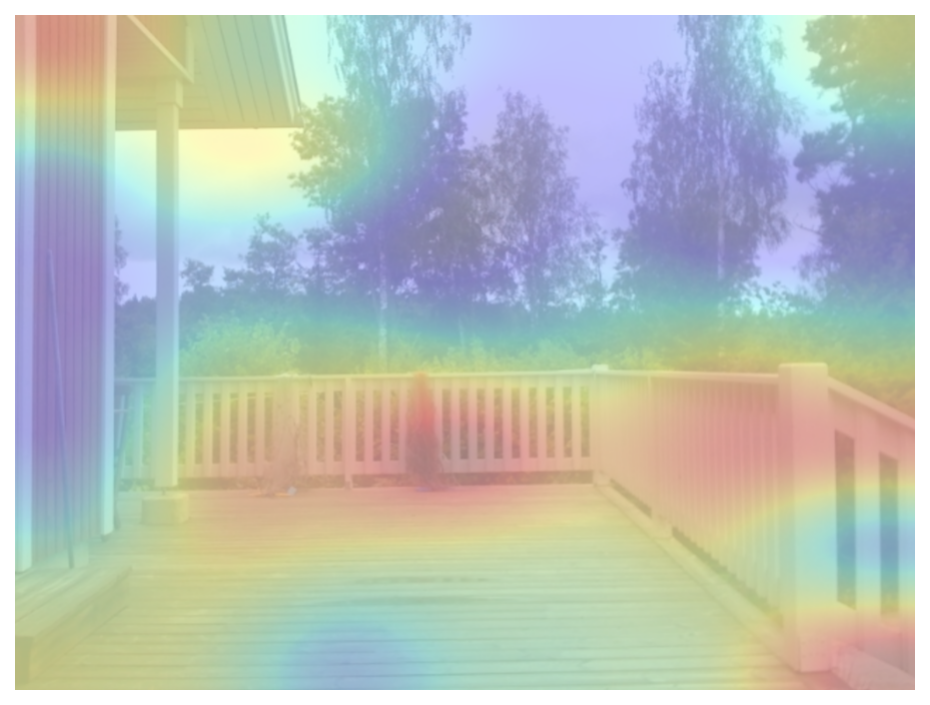} \\        
        Model (similarity) & Flex-ViT (0.601) & RoPE-ViT (0.563) & \method (0.628) \\
        \includegraphics[width=0.24\textwidth]{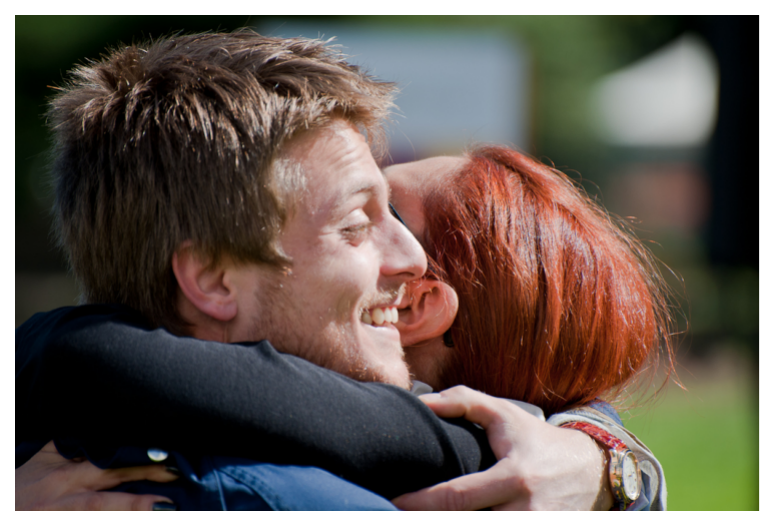} &
        \includegraphics[width=0.24\textwidth]{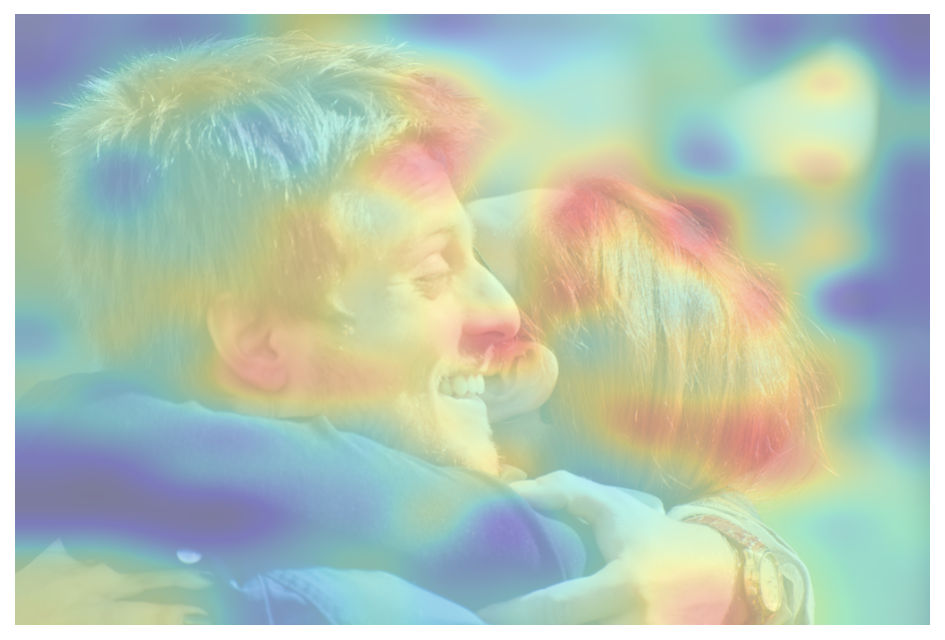} &
        \includegraphics[width=0.24\textwidth]{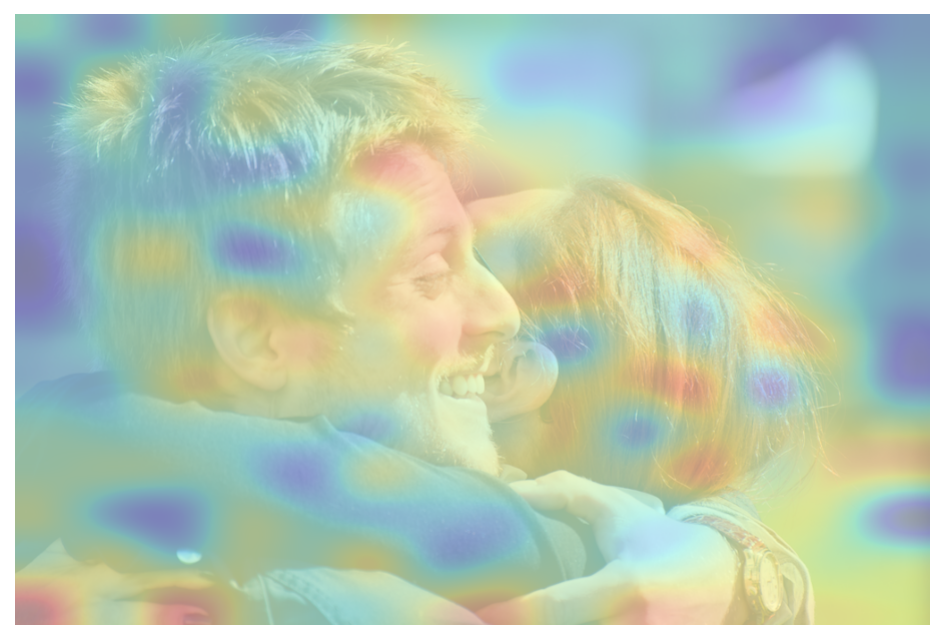} &
        \includegraphics[width=0.24\textwidth]{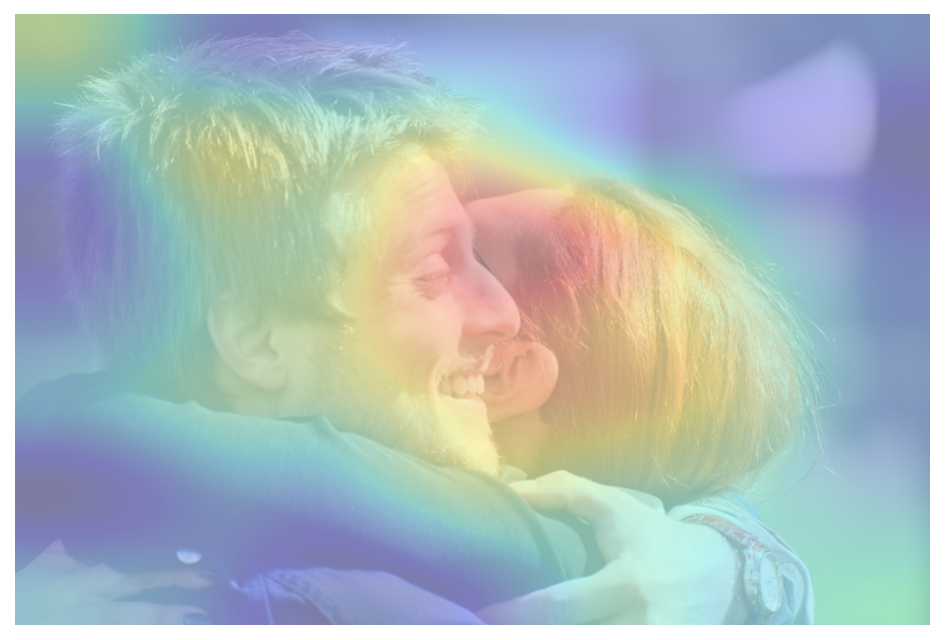}   \\      Model (similarity) & Flex-ViT (0.474) & RoPE-ViT (0.621) & \method (0.646) \\
        \includegraphics[width=0.24\textwidth]{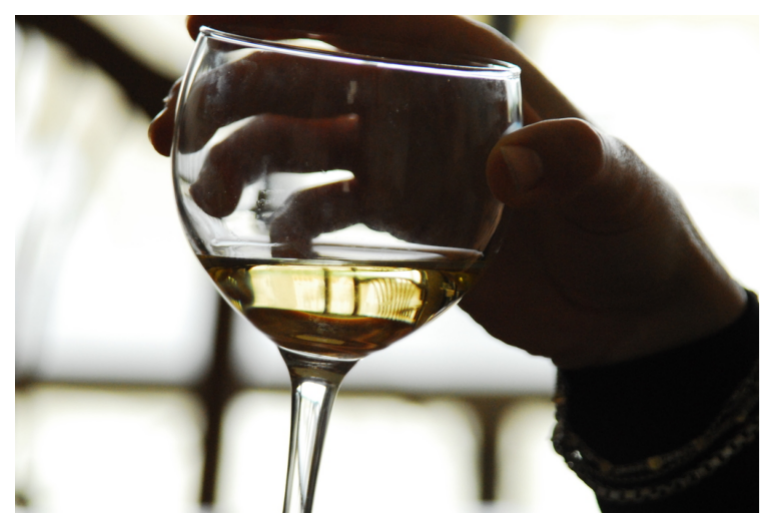} &
        \includegraphics[width=0.24\textwidth]{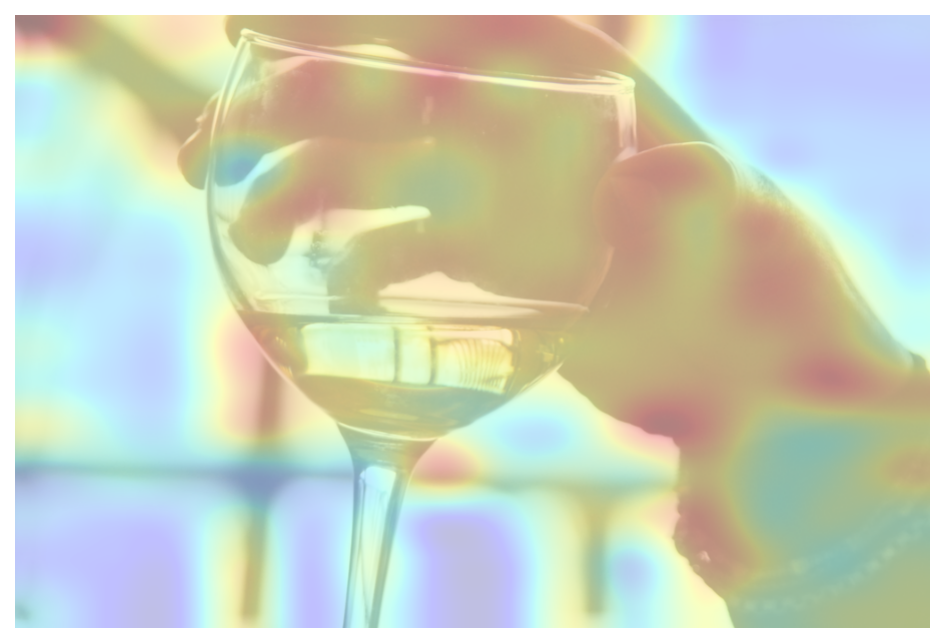} &
        \includegraphics[width=0.24\textwidth]{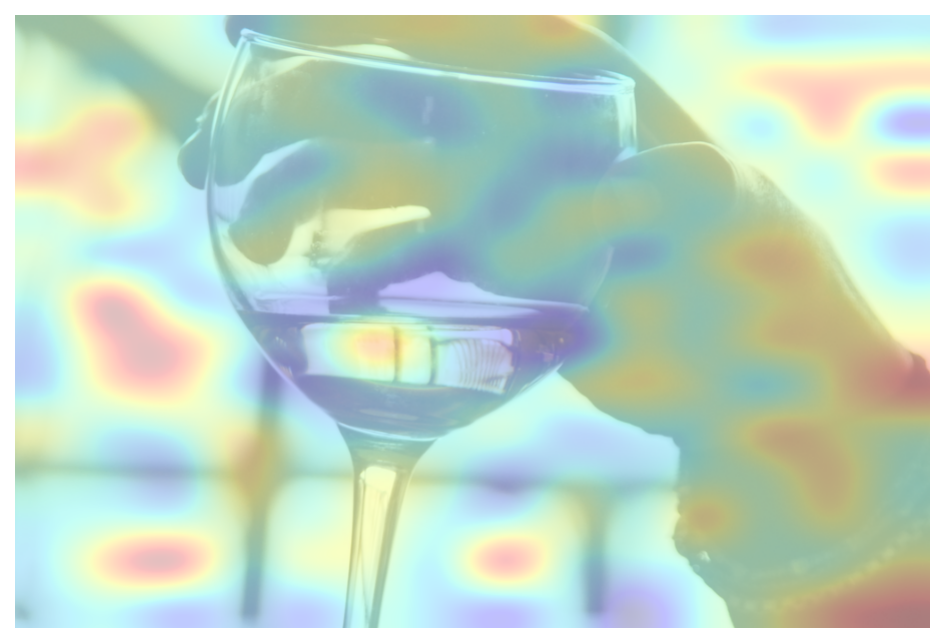} &
        \includegraphics[width=0.24\textwidth]{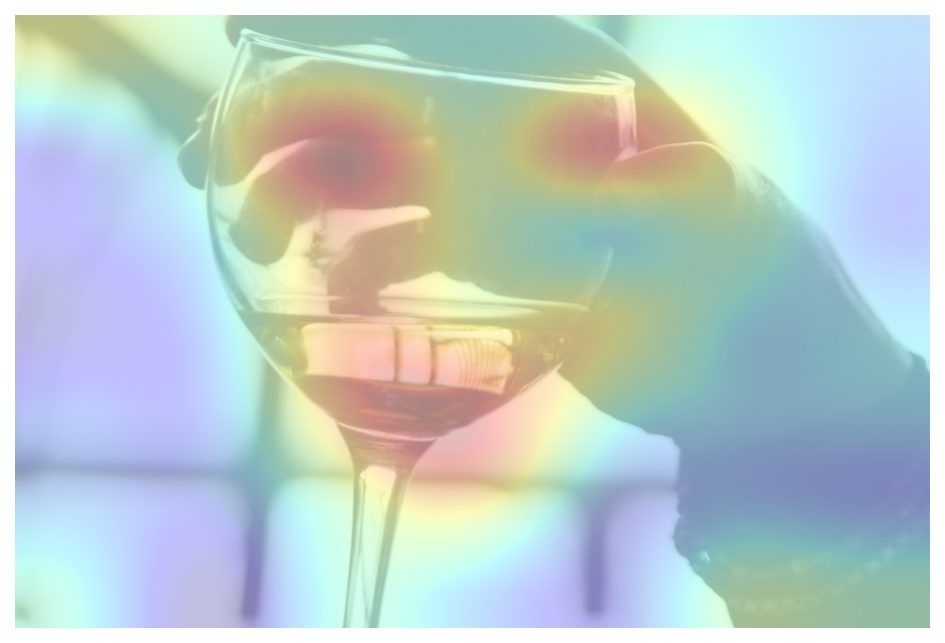}
    \end{tabular}
    \caption{Visualization of Image-Text similarity for the captions: "A large porch with a wooden fence and no roof." (top), ``A woman with red hair hugging a man.'' (middle) and ``A hand holding a glass that has some wine in it.'' (bottom). \method (similarity: 0.545) produces spatially coherent attention focused on the porch and fence, while RoPE-ViT (0.501) and FlexViT (0.452) show scattered, less interpretable patterns. Warmer colors indicate higher similarity.}
    \label{fig:nocaps_visuals}
    \vspace{-10pt}
\end{figure*}

%% file: tables/geometric_retrieval.tex
\begin{table}[t]
    \centering
    \begin{tabular}{ll c cc cc}
        \toprule
        & & & \multicolumn{2}{c}{\textbf{Uniformity} $\downarrow$} & \multicolumn{2}{c}{\textbf{Hubness} $\downarrow$} \\
        \cmidrule(lr){4-5} \cmidrule(lr){6-7}
        \textbf{Vision Tower} & \textbf{Text Tower} & \textbf{Alignment} $\downarrow$ & Image & Text & Image & Text \\
        \midrule
        FlexViT  & LLaMA   & 1.111 & -2.87 & -3.50 & 1.19 & 1.23 \\
        NaFlex   & LLaMA   & 1.039 & \textbf{-3.30} & -3.50 & \textbf{0.93} & 1.24 \\
        RoPE-ViT & LLaMA   & 1.052 & -3.16 & -3.52 & 1.04 & 1.25 \\
        \midrule
        \method~(VMamba)   & Mamba-1 & \textbf{0.982} & -3.18 & -3.54 & 1.01 & 1.15 \\
        \method~(VMamba)   & Mamba-2 & 1.000 & -3.16 & \textbf{-3.54} & 1.01 & \textbf{1.13} \\
        \bottomrule
    \end{tabular}%
    \caption{\textbf{Embedding geometry analysis on NoCaps.} We analyze alignment, uniformity, and hubness. \method achieves better alignment and lower text hubness than transformer baselines, explaining its superior retrieval performance. $\downarrow$ = lower is better for all metrics.}
    \label{tab:geometry}
    \vspace{-10pt}
\end{table}

%% file: figures/memory.tex
\begin{figure}[t]
\centering
\small
\begin{tabular}{cc}
\includegraphics[width=0.48\columnwidth]{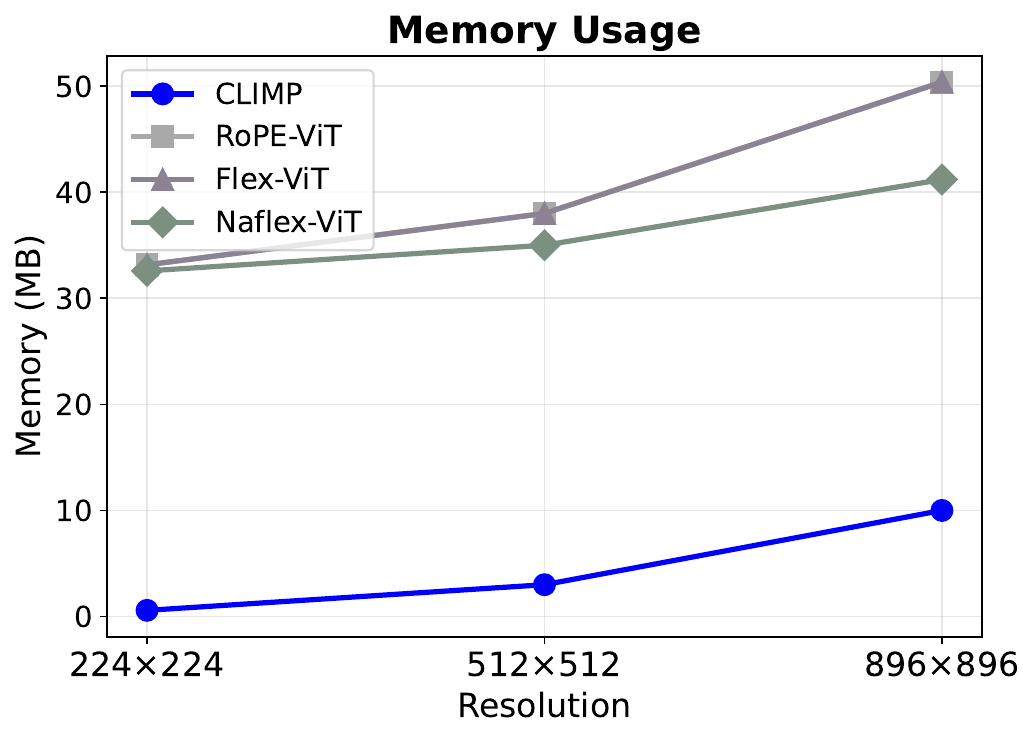} &
\includegraphics[width=0.48\columnwidth]{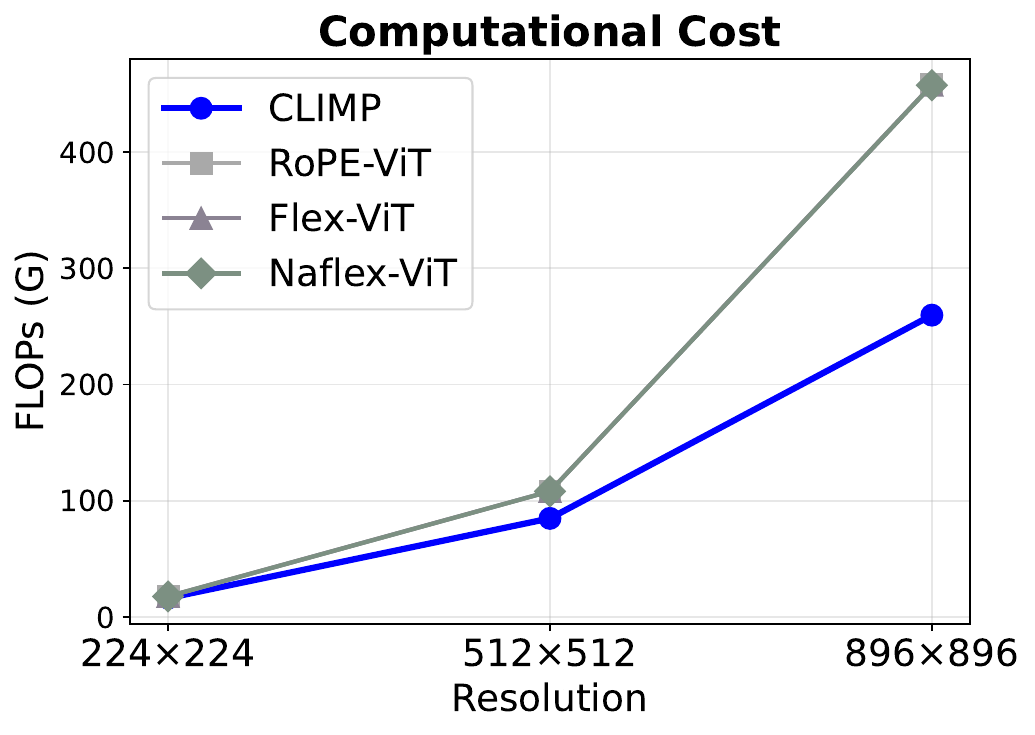} \\
\end{tabular}
\vspace{-4pt}
\caption{\textbf{Efficiency analysis.} \method achieves superior memory and computational efficiency across all resolutions. (Left) Memory overhead is 4--57$\times$ lower. (Right) FLOPs scale linearly, yielding up to 1.8$\times$ reduction—a gap that widens with resolution.
}
\label{fig:memory_flops}
\vspace{-6pt}
\end{figure}

%% file: tables/model_size.tex
\begin{table}[t]
\centering
\small
\begin{tabular}{lccc}
\toprule
\multirow{2}{*}{Params} & \method & FlexViT & RoPE-ViT \\
 & (Mamba2) & (LLaMA) & (LLaMA) \\
\midrule
22-30M  & 38.8 &  32.5 & 33.2\\
50M     & 42.8 & -- & -- \\
87M     & 43.5 & 38.3 & 40.7 \\
\bottomrule
\end{tabular}%
\caption{\textbf{Scaling behavior on ImageNet-1K zero-shot classification.} \method consistently outperforms ViT-based alternatives across all scales, with performance improving steadily as model size increases.}
\label{tab:model_size}
    \vspace{-10pt}
\end{table}

%% file: figures/scaling_laws.tex
\begin{figure}[t]
    \centering
    \small
    \includegraphics[width=0.5\linewidth]{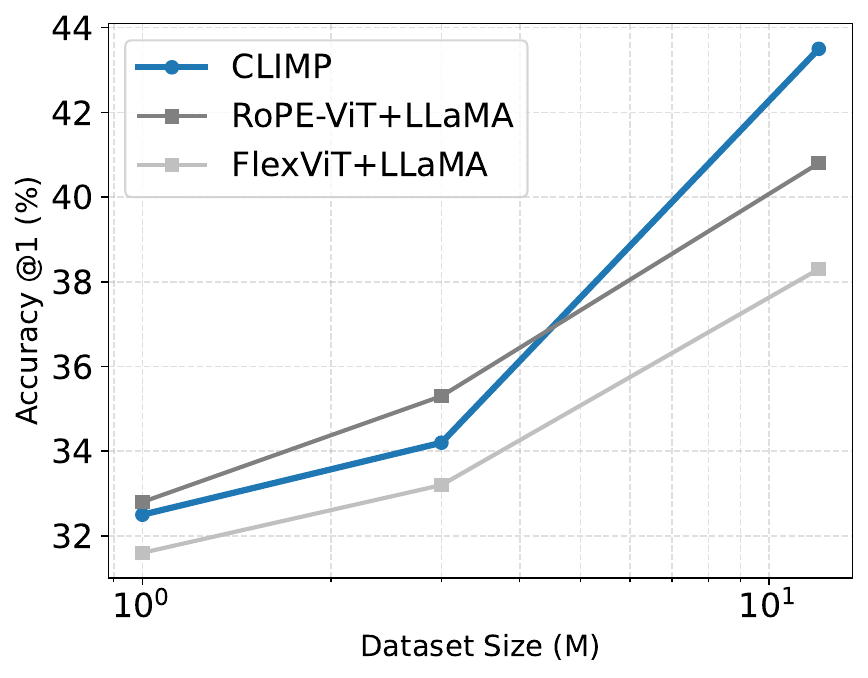}
    \vspace{-4pt}
    \caption{\method scaling laws trained on CC and evaluated on ImageNet-1K Acc@1. Performance improves consistently as training data scales from 1M to 12M samples, with no sign of saturation, indicating that our findings should scale further.}
    \label{fig:scaling_laws}
    \vspace{-10pt}
\end{figure}
% \vspace{-10pt}

%% file: tables/ablation_llm.tex
\begin{table}[t]
    \centering
    \begin{tabular}{ll cc}
        \toprule
        \textbf{Vision Tower} & \textbf{Text Tower} & \textbf{Acc@1/Acc@5} & \textbf{IR@5/TR@5} \\
        \midrule
        RoPE-ViT~\cite{heo2024ropevit} & LLaMA~\cite{touvron2023llama}   & 27.27/58.95 & 63.35/72.93 \\
        RoPE-ViT~\cite{heo2024ropevit} & Qwen2-1.5B~\cite{yang2024qwen2technicalreport} & 27.30/58.80 & 60.40/70.90 \\
        RoPE-ViT~\cite{heo2024ropevit} & Mamba-2~\cite{dao2024mamba2} & 27.42/58.69 & 62.59/72.99 \\
        RoPE-ViT~\cite{heo2024ropevit} & Mamba-1~\cite{gu2024mamba} & 28.10/58.54 & 63.17/74.28 \\
        RoPE-ViT~\cite{heo2024ropevit} & RoBERTa-L~\cite{liu2019roberta} & 27.47/58.64 & 64.82/74.72 \\
        RoPE-ViT~\cite{heo2024ropevit} & BERT-L~\cite{bert} &  27.80/58.84 & 63.93/74.15 \\
        \midrule
        VMamba~\cite{liu2024vmamba}   & LLaMA~\cite{touvron2023llama}   & 26.13/57.30 & 63.21/73.16 \\
        VMamba~\cite{liu2024vmamba}   & Hydra~\cite{hwang2024hydra}   & 28.63/56.98 & 61.24/71.35 \\
        VMamba~\cite{liu2024vmamba}   & Mamba-2~\cite{dao2024mamba2} & 26.94/\textbf{59.04} & 65.17/75.25 \\
        VMamba~\cite{liu2024vmamba}   & Mamba-1~\cite{gu2024mamba} & \textbf{29.59}/58.53 & \textbf{65.54}/\textbf{77.03} \\
        \bottomrule
    \end{tabular}%
    \caption{\textbf{Architectural synergy.} We compare language model backbones across vision encoders. While RoPE-ViT shows minimal sensitivity to text encoder choice, VMamba consistently benefits from Mamba-based text encoders, suggesting that matched SSMs learn more compatible cross-modal representations. Hydra~\cite{hwang2024hydra}, a bidirectional SSM, underperforms the autoregressive Mamba LLMs. }
    \label{tab:llm_ablation}
    \vspace{-16pt}
\end{table}
% \vspace{-10pt}

%% file: conclusions_limitations.tex
\section{Conclusions}
\label{sec:conclusions}
We introduced \method, the first contrastive vision-language model built entirely on Mamba. By replacing ViT with VMamba and pairing it with Mamba LLMs, \method achieves linear complexity in both modalities while producing superior representation quality and robustness.
Our experiments reveal: (1) superior retrieval performance, driven by tighter cross-modal alignment and reduced hubness; (2) improved OOD robustness, notably surpassing CLIP-ViT-B/16 trained on LAION-2B on ImageNet-O; (3) memory and FLOPs reductions at high resolutions with native variable-resolution support; (4) dense captioning retrieval beyond CLIP's token limit; (5) spatial inductive bias contributing to sample-efficient learning; and (6) favorable scaling behavior across model and dataset sizes, with unsaturated performance curves indicating that these advantages are architectural and should persist at larger scales.

These findings establish SSMs as a viable alternative to transformers for vision-language pre-training, overcoming limitations of CLIP implementations for high-resolution, memory-constrained, and long-context applications. Future directions include scaling to larger models and datasets, extending to generative vision-language tasks, and developing hybrid architectures~\cite{nvidia_nemotron_nano_v3_2025}.

\section{Acknowledgments}
\label{sec:acks}
This work was partially supported by Tel Aviv University Center for AI and Data Science (TAD) and Blavatnik Interdisciplinary Cyber Research Center (ICRC).

%% file: custom.bib
@inproceedings{radford2021learning,
  title={Learning transferable visual models from natural language supervision},
  author={Radford, Alec and Kim, Jong Wook and Hallacy, Chris and Ramesh, Aditya and Goh, Gabriel and Agarwal, Sandhini and Sastry, Girish and Askell, Amanda and Mishkin, Pamela and Clark, Jack and others},
  booktitle={International conference on machine learning},
  pages={8748--8763},
  year={2021},
  organization={PmLR}
}

@inproceedings{dosovitskiy2021image,
  title={An Image is Worth 16x16 Words: Transformers for Image Recognition at Scale},
  author={Dosovitskiy, Alexey and Beyer, Lucas and Kolesnikov, Alexander and Weissenborn, Dirk and Zhai, Xiaohua and Unterthiner, Thomas and Dehghani, Mostafa and Minderer, Matthias and Heigold, Georg and Gelly, Sylvain and Uszkoreit, Jakob and Houlsby, Neil},
  booktitle={International Conference on Learning Representations},
  year={2021},
}

@article{gu2024mamba,
  title={Mamba: Linear-time sequence modeling with selective state spaces},
  author={Gu, Albert and Dao, Tri},
  journal={arXiv preprint arXiv:2312.00752},
  year={2023}
}

@inproceedings{zhu2024vision,
  title={Vision Mamba: Efficient Visual Representation Learning with Bidirectional State Space Model},
  author={Zhu, Lianghui and Liao, Bencheng and Zhang, Qian and Wang, Xinlong and Liu, Wenyu and Wang, Xinggang},
  booktitle={International Conference on Machine Learning},
  pages={62429--62442},
  year={2024},
  organization={PMLR}
}

@article{liu2024vmamba,
  title={Vmamba: Visual state space model},
  author={Liu, Yue and Tian, Yunjie and Zhao, Yuzhong and Yu, Hongtian and Xie, Lingxi and Wang, Yaowei and Ye, Qixiang and Jiao, Jianbin and Liu, Yunfan},
  journal={Advances in neural information processing systems},
  volume={37},
  pages={103031--103063},
  year={2024}
}

@article{du2024understanding,
  title={Understanding robustness of visual state space models for image classification},
  author={Du, Chengbin and Li, Yanxi and Xu, Chang},
  journal={arXiv preprint arXiv:2403.10935},
  year={2024}
}

@article{huang2024clipmamba,
  title={Clip-mamba: Clip pretrained mamba models with ood and hessian evaluation},
  author={Huang, Weiquan and Shen, Yifei and Yang, Yifan},
  journal={arXiv preprint arXiv:2404.19394},
  year={2024}
}

@article{su2024roformer,
  title={Roformer: Enhanced transformer with rotary position embedding},
  author={Su, Jianlin and Ahmed, Murtadha and Lu, Yu and Pan, Shengfeng and Bo, Wen and Liu, Yunfeng},
  journal={Neurocomputing},
  volume={568},
  pages={127063},
  year={2024},
  publisher={Elsevier}
}

@article{touvron2023llama,
  title={The llama 3 herd of models},
  author={Grattafiori, Aaron and Dubey, Abhimanyu and Jauhri, Abhinav and Pandey, Abhinav and Kadian, Abhishek and Al-Dahle, Ahmad and Letman, Aiesha and Mathur, Akhil and Schelten, Alan and Vaughan, Alex and others},
  journal={arXiv preprint arXiv:2407.21783},
  year={2024}
}

@inproceedings{dao2024mamba2,
  title={Transformers are SSMs: generalized models and efficient algorithms through structured state space duality},
  author={Dao, Tri and Gu, Albert},
  booktitle={Proceedings of the 41st International Conference on Machine Learning},
  pages={10041--10071},
  year={2024}
}

@inproceedings{cherti2023reproducible,
  title={Reproducible scaling laws for contrastive language-image learning},
  author={Cherti, Mehdi and Beaumont, Romain and Wightman, Ross and Wortsman, Mitchell and Ilharco, Gabriel and Gordon, Cade and Schuhmann, Christoph and Schmidt, Ludwig and Jitsev, Jenia},
  booktitle={Proceedings of the IEEE/CVF conference on computer vision and pattern recognition},
  pages={2818--2829},
  year={2023}
}

@inproceedings{zhai2023sigmoid,
  title={Sigmoid loss for language image pre-training},
  author={Zhai, Xiaohua and Mustafa, Basil and Kolesnikov, Alexander and Beyer, Lucas},
  booktitle={Proceedings of the IEEE/CVF international conference on computer vision},
  pages={11975--11986},
  year={2023}
}

@article{sun2023evaclip,
  title={Eva-clip: Improved training techniques for clip at scale},
  author={Sun, Quan and Fang, Yuxin and Wu, Ledell and Wang, Xinlong and Cao, Yue},
  journal={arXiv preprint arXiv:2303.15389},
  year={2023}
}

@inproceedings{xu2024demystifying,
  title={Demystifying clip data},
  author={Xu, Hu and Xie, Saining and Tan, Xiaoqing and Huang, Po-Yao and Howes, Russell and Sharma, Vasu and Li, Shang-Wen and Ghosh, Gargi and Zettlemoyer, Luke and Feichtenhofer, Christoph},
  booktitle={International Conference on Learning Representations},
  year={2024}
}

@article{tschannen2025siglip2,
  title={Siglip 2: Multilingual vision-language encoders with improved semantic understanding, localization, and dense features},
  author={Tschannen, Michael and Gritsenko, Alexey and Wang, Xiao and Naeem, Muhammad Ferjad and Alabdulmohsin, Ibrahim and Parthasarathy, Nikhil and Evans, Talfan and Beyer, Lucas and Xia, Ye and Mustafa, Basil and others},
  journal={arXiv preprint arXiv:2502.14786},
  year={2025}
}

@article{taori2020measuring,
  title={Measuring robustness to natural distribution shifts in image classification},
  author={Taori, Rohan and Dave, Achal and Shankar, Vaishaal and Carlini, Nicholas and Recht, Benjamin and Schmidt, Ludwig},
  journal={Advances in Neural Information Processing Systems},
  volume={33},
  pages={18583--18599},
  year={2020}
}

@inproceedings{fang2022data,
  title={Data determines distributional robustness in contrastive language image pre-training (clip)},
  author={Fang, Alex and Ilharco, Gabriel and Wortsman, Mitchell and Wan, Yuhao and Shankar, Vaishaal and Dave, Achal and Schmidt, Ludwig},
  booktitle={International conference on machine learning},
  pages={6216--6234},
  year={2022},
  organization={PMLR}
}

@article{wang2024sober,
  title={A sober look at the robustness of clips to spurious features},
  author={Wang, Qizhou and Lin, Yong and Chen, Yongqiang and Schmidt, Ludwig and Han, Bo and Zhang, Tong},
  journal={Advances in Neural Information Processing Systems},
  volume={37},
  pages={122484--122523},
  year={2024}
}

@article{ruan2024vmunet,
  title={Vm-unet: Vision mamba unet for medical image segmentation},
  author={Ruan, Jiacheng and Li, Jincheng and Xiang, Suncheng},
  journal={ACM Transactions on Multimedia Computing, Communications and Applications},
  year={2024},
  publisher={ACM New York, NY}
}

@inproceedings{li2024videomamba,
  title={Videomamba: State space model for efficient video understanding},
  author={Li, Kunchang and Li, Xinhao and Wang, Yi and He, Yinan and Wang, Yali and Wang, Limin and Qiao, Yu},
  booktitle={European conference on computer vision},
  pages={237--255},
  year={2024},
  organization={Springer}
}

@article{liu2024point,
  title={Point mamba: A novel point cloud backbone based on state space model with octree-based ordering strategy},
  author={Liu, Jiuming and Yu, Ruiji and Wang, Yian and Zheng, Yu and Deng, Tianchen and Ye, Weicai and Wang, Hesheng},
  journal={arXiv preprint arXiv:2403.06467},
  year={2024}
}

@article{laion400m,
  title={Laion-400m: Open dataset of clip-filtered 400 million image-text pairs},
  author={Schuhmann, Christoph and Vencu, Richard and Beaumont, Romain and Kaczmarczyk, Robert and Mullis, Clayton and Katta, Aarush and Coombes, Theo and Jitsev, Jenia and Komatsuzaki, Aran},
  journal={arXiv preprint arXiv:2111.02114},
  year={2021}
}

@article{schuhmann2022laion,
  title={Laion-5b: An open large-scale dataset for training next generation image-text models},
  author={Schuhmann, Christoph and Beaumont, Romain and Vencu, Richard and Gordon, Cade and Wightman, Ross and Cherti, Mehdi and Coombes, Theo and Katta, Aarush and Mullis, Clayton and Wortsman, Mitchell and others},
  journal={Advances in neural information processing systems},
  volume={35},
  pages={25278--25294},
  year={2022}
}

@inproceedings{heo2024ropevit,
  title={Rotary position embedding for vision transformer},
  author={Heo, Byeongho and Park, Song and Han, Dongyoon and Yun, Sangdoo},
  booktitle={European Conference on Computer Vision},
  pages={289--305},
  year={2024},
  organization={Springer}
}

@inproceedings{beyer2023flexvit,
  title={Flexivit: One model for all patch sizes},
  author={Beyer, Lucas and Izmailov, Pavel and Kolesnikov, Alexander and Caron, Mathilde and Kornblith, Simon and Zhai, Xiaohua and Minderer, Matthias and Tschannen, Michael and Alabdulmohsin, Ibrahim and Pavetic, Filip},
  booktitle={Proceedings of the IEEE/CVF Conference on Computer Vision and Pattern Recognition},
  pages={14496--14506},
  year={2023}
}

@article{dehghani2023patch,
  title={Patch n’pack: Navit, a vision transformer for any aspect ratio and resolution},
  author={Dehghani, Mostafa and Mustafa, Basil and Djolonga, Josip and Heek, Jonathan and Minderer, Matthias and Caron, Mathilde and Steiner, Andreas and Puigcerver, Joan and Geirhos, Robert and Alabdulmohsin, Ibrahim M and others},
  journal={Advances in Neural Information Processing Systems},
  volume={36},
  pages={2252--2274},
  year={2023}
}

@inproceedings{cc12m,
  title={Conceptual 12m: Pushing web-scale image-text pre-training to recognize long-tail visual concepts},
  author={Changpinyo, Soravit and Sharma, Piyush and Ding, Nan and Soricut, Radu},
  booktitle={Proceedings of the IEEE/CVF conference on computer vision and pattern recognition},
  pages={3558--3568},
  year={2021}
}

@inproceedings{deng2009imagenet,
  title={Imagenet: A large-scale hierarchical image database},
  author={Deng, Jia and Dong, Wei and Socher, Richard and Li, Li-Jia and Li, Kai and Fei-Fei, Li},
  booktitle={2009 IEEE conference on computer vision and pattern recognition},
  pages={248--255},
  year={2009},
  organization={Ieee}
}

@article{patro2024simba,
  title={Simba: Simplified mamba-based architecture for vision and multivariate time series},
  author={Patro, Badri N and Agneeswaran, Vijay S},
  journal={arXiv preprint arXiv:2403.15360},
  year={2024}
}

@misc{clip_benchmarks,
  author       = {Cherti, Mehdi and
                  Beaumont, Romain},
  title        = {CLIP benchmark},
  month        = nov,
  year         = 2022,
  publisher    = {Zenodo},
  doi          = {10.5281/zenodo.15403103},
  url          = {https://doi.org/10.5281/zenodo.15403103},
  swhid        = {swh:1:dir:8cf49a5dd06f59224844a1e767337a1d14ee56c2
                   ;origin=https://doi.org/10.5281/zenodo.15403102;vi
                   sit=swh:1:snp:dd153b26f702d614346bf814f723d59fef3d
                   77a2;anchor=swh:1:rel:cff2aeb98f42583b44fdab5374e9
                   fa71793f2cff;path=CLIP\_benchmark-main
                  },
}

@inproceedings{agrawal2019nocaps,
  title={Nocaps: Novel object captioning at scale},
  author={Agrawal, Harsh and Desai, Karan and Wang, Yufei and Chen, Xinlei and Jain, Rishabh and Johnson, Mark and Batra, Dhruv and Parikh, Devi and Lee, Stefan and Anderson, Peter},
  booktitle={Proceedings of the IEEE/CVF international conference on computer vision},
  pages={8948--8957},
  year={2019}
}

@inproceedings{ThapliyalCrossmodal2022,
  title={Crossmodal-3600: A massively multilingual multimodal evaluation dataset},
  author={Thapliyal, Ashish V and Tuset, Jordi Pont and Chen, Xi and Soricut, Radu},
  booktitle={Proceedings of the 2022 Conference on Empirical Methods in Natural Language Processing},
  pages={715--729},
  year={2022}
}

@inproceedings{imagenet_ao,
  title={Natural adversarial examples. 2021 IEEE},
  author={Hendrycks, Dan and Zhao, Kevin and Basart, Steven and Steinhardt, Jacob and Song, Dawn Xiaodong},
  booktitle={CVF Conference on Computer Vision and Pattern Recognition (CVPR)},
  pages={15257--15266},
  year={2019}
}

@inproceedings{imagenet_r,
  title={The many faces of robustness: A critical analysis of out-of-distribution generalization},
  author={Hendrycks, Dan and Basart, Steven and Mu, Norman and Kadavath, Saurav and Wang, Frank and Dorundo, Evan and Desai, Rahul and Zhu, Tyler and Parajuli, Samyak and Guo, Mike and others},
  booktitle={Proceedings of the IEEE/CVF international conference on computer vision},
  pages={8340--8349},
  year={2021}
}

@article{imagenet_sketch,
  title={Learning robust global representations by penalizing local predictive power},
  author={Wang, Haohan and Ge, Songwei and Lipton, Zachary and Xing, Eric P},
  journal={Advances in neural information processing systems},
  volume={32},
  year={2019}
}

@inproceedings{imagenetv2,
  title={Do imagenet classifiers generalize to imagenet?},
  author={Recht, Benjamin and Roelofs, Rebecca and Schmidt, Ludwig and Shankar, Vaishaal},
  booktitle={International conference on machine learning},
  pages={5389--5400},
  year={2019},
  organization={PMLR}
}

@article{nvidia_nemotron_nano_v3_2025,
  title={Nemotron 3 Nano: Open, Efficient Mixture-of-Experts Hybrid Mamba-Transformer Model for Agentic Reasoning},
  author={Blakeman, Aaron and Grattafiori, Aaron and Basant, Aarti and Gupta, Abhibha and Khattar, Abhinav and Renduchintala, Adi and Vavre, Aditya and Shukla, Akanksha and Bercovich, Akhiad and Ficek, Aleksander and others},
  journal={arXiv preprint arXiv:2512.20848},
  year={2025}
}

@article{tamayo2025spuriousvit,
  title={Your attention matters: to improve model robustness to noise and spurious correlations},
  author={Tamayo-Rousseau, Camilo and Zhao, Yunjia and Zhang, Yiqun and Balestriero, Randall},
  journal={arXiv preprint arXiv:2507.20453},
  year={2025}
}

@inproceedings{zhou2025fighting,
  title={Fighting spurious correlations in text classification via a causal learning perspective},
  author={Zhou, Yuqing and Zhu, Ziwei},
  booktitle={Proceedings of the 2025 Conference of the Nations of the Americas Chapter of the Association for Computational Linguistics: Human Language Technologies (Volume 1: Long Papers)},
  pages={4264--4274},
  year={2025}
}

@inproceedings{OnoeDocci2024,
  author        = {Yasumasa Onoe and Sunayana Rane and Zachary Berger and Yonatan Bitton and Jaemin Cho and Roopal Garg and
    Alexander Ku and Zarana Parekh and Jordi Pont-Tuset and Garrett Tanzer and Su Wang and Jason Baldridge},
  title         = {{DOCCI: Descriptions of Connected and Contrasting Images}},
  booktitle     = {ECCV},
  year          = {2024}
}

@inproceedings{bert,
  title={Bert: Pre-training of deep bidirectional transformers for language understanding},
  author={Devlin, Jacob and Chang, Ming-Wei and Lee, Kenton and Toutanova, Kristina},
  booktitle={Proceedings of the 2019 conference of the North American chapter of the association for computational linguistics: human language technologies, volume 1 (long and short papers)},
  pages={4171--4186},
  year={2019}
}

@article{radovanovic2010hubs,
  title={Hubs in space: Popular nearest neighbors in high-dimensional data},
  author={Radovanovic, Milos and Nanopoulos, Alexandros and Ivanovic, Mirjana},
  journal={Journal of machine learning research},
  volume={11},
  number={sept},
  pages={2487--2531},
  year={2010}
}

@inproceedings{wang2020understanding,
  title={Understanding contrastive representation learning through alignment and uniformity on the hypersphere},
  author={Wang, Tongzhou and Isola, Phillip},
  booktitle={International conference on machine learning},
  pages={9929--9939},
  year={2020},
  organization={PMLR}
}

@article{zhang2024adversarial,
  title={Adversarial hubness in multi-modal retrieval},
  author={Zhang, Tingwei and Suya, Fnu and Jha, Rishi and Zhang, Collin and Shmatikov, Vitaly},
  journal={arXiv preprint arXiv:2412.14113},
  year={2024}
}

@inproceedings{dascoli2021convit,
  title={Convit: Improving vision transformers with soft convolutional inductive biases},
  author={d’Ascoli, St{\'e}phane and Touvron, Hugo and Leavitt, Matthew L and Morcos, Ari S and Biroli, Giulio and Sagun, Levent},
  booktitle={International conference on machine learning},
  pages={2286--2296},
  year={2021},
  organization={PMLR}
}

@misc{krizhevsky2009cifar,
  title={Learning multiple layers of features from tiny images},
  author={Krizhevsky, Alex and Hinton, Geoffrey and others},
  year={2009}
}

@article{hwang2024hydra,
  title={Hydra: Bidirectional state space models through generalized matrix mixers},
  author={Hwang, Sukjun and Lahoti, Aakash and Puduppully, Ratish and Dao, Tri and Gu, Albert},
  journal={Advances in Neural Information Processing Systems},
  volume={37},
  pages={110876--110908},
  year={2024}
}

@article{liu2019roberta,
  title={Roberta: A robustly optimized bert pretraining approach},
  author={Liu, Yinhan and Ott, Myle and Goyal, Naman and Du, Jingfei and Joshi, Mandar and Chen, Danqi and Levy, Omer and Lewis, Mike and Zettlemoyer, Luke and Stoyanov, Veselin},
  journal={arXiv preprint arXiv:1907.11692},
  year={2019}
}

@inproceedings{ALIGN,
  title={Scaling up visual and vision-language representation learning with noisy text supervision},
  author={Jia, Chao and Yang, Yinfei and Xia, Ye and Chen, Yi-Ting and Parekh, Zarana and Pham, Hieu and Le, Quoc and Sung, Yun-Hsuan and Li, Zhen and Duerig, Tom},
  booktitle={International conference on machine learning},
  pages={4904--4916},
  year={2021},
  organization={PMLR}
}

@inproceedings{malik2025towards,
  title={Towards evaluating the robustness of visual state space models},
  author={Malik, Hashmat Shadab and Shamshad, Fahad and Naseer, Muzammal and Nandakumar, Karthik and Khan, Fahad Shahbaz and Khan, Salman},
  booktitle={Proceedings of the Computer Vision and Pattern Recognition Conference},
  pages={3544--3553},
  year={2025}
}

@inproceedings{zimerman2024viewing,
  title={Viewing transformers through the lens of long convolutions layers},
  author={Zimerman, Itamar and Wolf, Lior},
  booktitle={Forty-first International Conference on Machine Learning},
  year={2024}
}

@article{liu-etal-2025-data-language,
  title={Data or Language Supervision: What Makes CLIP Better than DINO?},
  author={Liu, Yiming and Zhang, Yuhui and Ghosh, Dhruba and Schmidt, Ludwig and Yeung-Levy, Serena},
  journal={arXiv preprint arXiv:2510.11835},
  year={2025}
}

@article{qiao2025univitar,
  title={UniViTAR: Unified Vision Transformer with Native Resolution},
  author={Limeng Qiao and Yiyang Gan and Bairui Wang and Jie Qin and Shuang Xu and Siqi Yang and Lin Ma},
  journal={ArXiv},
  year={2025},
}

@article{huang2024llm2clip,
  title={Llm2clip: Powerful language model unlocks richer visual representation},
  author={Huang, Weiquan and Wu, Aoqi and Yang, Yifan and Luo, Xufang and Yang, Yuqing and Hu, Liang and Dai, Qi and Wang, Chunyu and Dai, Xiyang and Chen, Dongdong and others},
  journal={arXiv preprint arXiv:2411.04997},
  year={2024}
}

@article{He2015DeepRL,
  title={Deep Residual Learning for Image Recognition},
  author={Kaiming He and X. Zhang and Shaoqing Ren and Jian Sun},
  journal={2016 IEEE Conference on Computer Vision and Pattern Recognition (CVPR)},    
}

@misc{open_clip,
  author       = {Ilharco, Gabriel and
                  Wortsman, Mitchell and
                  Wightman, Ross and
                  Gordon, Cade and
                  Carlini, Nicholas and
                  Taori, Rohan and
                  Dave, Achal and
                  Shankar, Vaishaal and
                  Namkoong, Hongseok and
                  Miller, John and
                  Hajishirzi, Hannaneh and
                  Farhadi, Ali and
                  Schmidt, Ludwig},
  title        = {OpenCLIP},
  month        = jul,
  year         = 2021,
  note         = {If you use this software, please cite it as below.},
  publisher    = {Zenodo},
  version      = {0.1},
  doi          = {10.5281/zenodo.5143773},
  url          = {https://doi.org/10.5281/zenodo.5143773}
}

@ARTICLE{Yellinek20253VL,
  author={Yellinek, Nir and Karlinsky, Leonid and Giryes, Raja},
  journal={IEEE Transactions on Image Processing}, 
  title={3VL: Using Trees to Improve Vision-Language Models’ Interpretability}, 
  year={2025},
  volume={34},
  pages={495-509},
}

@inproceedings{Ben-Kish2025DeciMamba,
 author = {Ben-Kish, Assaf and Zimerman, Itamar and Abu-Hussein, Shady and Cohen, Nadav and Globerson, Amir and Wolf, Lior and Giryes, Raja},
 booktitle = {International Conference on Learning Representations},
 pages = {101148--101170},
 title = {DeciMamba: Exploring the Length Extrapolation Potential of Mamba},
 year = {2025},
}

@inproceedings{ben-kish2025overflow,
title={Overflow Prevention Enhances Long-Context Recurrent {LLM}s},
author={Assaf Ben-Kish and Itamar Zimerman and Muhammad Jehanzeb Mirza and Lior Wolf and James R. Glass and Leonid Karlinsky and Raja Giryes},
booktitle={Conference on Language Modeling},
year={2025},
}

@inproceedings{lu2025mamba,
      title={Mamba Modulation: On the Length Generalization of Mamba Models},
      author={Peng Lu and Jerry Huang and Qiuhao Zeng and Xinyu Wang and Boxing Chen and Philippe Langlais and Yufei Cui},
      booktitle={Conference on Neural Information Processing Systems (NeurIPS)},
      year={2025},
}

@inproceedings{bar2025revisiting,
title={Revisiting Glorot Initialization for Long-Range Linear Recurrences},
author={Noga Bar and Mariia Seleznova and Yotam Alexander and Gitta Kutyniok and Raja Giryes},
booktitle={Conference on Neural Information Processing Systems (NeurIPS)},
year={2025},
}

@inproceedings{yanuka2025bridging,
    title = "Bridging the Visual Gap: Fine-Tuning Multimodal Models with Knowledge-Adapted Captions",
    author = "Yanuka, Moran  and
      Ben-Kish, Assaf  and
      Bitton, Yonatan  and
      Szpektor, Idan  and
      Giryes, Raja",
    booktitle = "Conference of the Nations of the Americas Chapter of the Association for Computational Linguistics (NAACL)",
    year = "2025",
    pages = "10497--10518",
}

@misc{yang2024qwen2technicalreport,
      title={Qwen2 Technical Report}, 
      author={An Yang and Baosong Yang and Binyuan Hui and Bo Zheng and Bowen Yu and Chang Zhou and Chengpeng Li and Chengyuan Li and Dayiheng Liu and Fei Huang and Guanting Dong and Haoran Wei and Huan Lin and Jialong Tang and Jialin Wang and Jian Yang and Jianhong Tu and Jianwei Zhang and Jianxin Ma and Jianxin Yang and Jin Xu and Jingren Zhou and Jinze Bai and Jinzheng He and Junyang Lin and Kai Dang and Keming Lu and Keqin Chen and Kexin Yang and Mei Li and Mingfeng Xue and Na Ni and Pei Zhang and Peng Wang and Ru Peng and Rui Men and Ruize Gao and Runji Lin and Shijie Wang and Shuai Bai and Sinan Tan and Tianhang Zhu and Tianhao Li and Tianyu Liu and Wenbin Ge and Xiaodong Deng and Xiaohuan Zhou and Xingzhang Ren and Xinyu Zhang and Xipin Wei and Xuancheng Ren and Xuejing Liu and Yang Fan and Yang Yao and Yichang Zhang and Yu Wan and Yunfei Chu and Yuqiong Liu and Zeyu Cui and Zhenru Zhang and Zhifang Guo and Zhihao Fan},
      year={2024},
      eprint={2407.10671},
      archivePrefix={arXiv},
      primaryClass={cs.CL},
      url={https://arxiv.org/abs/2407.10671}, 
}
